\documentclass[preprint]{elsarticle}

\usepackage{hyperref}
\usepackage{graphicx}
\usepackage{amsmath}
\usepackage{amssymb}
\usepackage{subfigure}
\usepackage{multirow}
\usepackage{color}
\usepackage{bbding}
\usepackage{url}
\usepackage{adjustbox}

\newcommand{\etal}{\textit{et al.~}}
\newcommand{\ie}{\textit{i.e.}}
\newcommand{\eg}{\textit{e.g.}}

\journal{Journal of Information Sciences}

\begin{document}

\begin{frontmatter}

\title{Progressive Perception-Oriented Network \\ for Single Image Super-Resolution}

%% Group authors per affiliation:
\author[Xidian]{Zheng~Hui}
\ead{zheng\_hui@aliyun.com}

\author[Xidian]{Jie~Li}
\ead{leejie@mail.xidian.edu.cn}

\author[Xidian,Chongqing]{Xinbo~Gao\corref{mycorrespondingauthor}}
\ead{xbgao@mail.xidian.edu.cn}
\cortext[mycorrespondingauthor]{Corresponding author}

\author[Xidian]{Xiumei~Wang}
\ead{wangxm@xidian.edu.cn}

\address[Xidian]{Video \& Image Processing System (VIPS) Lab, School of Electronic Engineering, Xidian University, No.2, South Taibai Road, Xi'an 710071, China}
\address[Chongqing]{The Chongqing Key Laboratory of Image Cognition, Chongqing University of Posts and Telecommunications, Chongqing 400065, China}

\begin{abstract}
Recently, it has been demonstrated that deep neural networks can significantly improve the performance of single image super-resolution (SISR). Numerous studies have concentrated on raising the quantitative quality of super-resolved (SR) images. However, these methods that target PSNR maximization usually produce blurred images at large upscaling factor. The introduction of generative adversarial networks (GANs) can mitigate this issue and show impressive results with synthetic high-frequency textures. Nevertheless, these GAN-based approaches always have a tendency to add fake textures and even artifacts to make the SR image of visually higher-resolution. In this paper, we propose a novel perceptual image super-resolution method that progressively generates visually high-quality results by constructing a stage-wise network. Specifically, the first phase concentrates on minimizing pixel-wise error, and the second stage utilizes the features extracted by the previous stage to pursue results with better structural retention. The final stage employs fine structure features distilled by the second phase to produce more realistic results. In this way, we can maintain the pixel, and structural level information in the perceptual image as much as possible. It is useful to note that the proposed method can build three types of images in a feed-forward process. Also, we explore a new generator that adopts multi-scale hierarchical features fusion. Extensive experiments on benchmark datasets show that our approach is superior to the state-of-the-art methods. Code is available at \url{https://github.com/Zheng222/PPON}.
\end{abstract}

\begin{keyword}
Perceptual image super-resolution\sep progressive related works learning\sep multi-scale hierarchical fusion
\end{keyword}

\end{frontmatter}

%\linenumbers

\section{Introduction}

Due to the emergence of deep learning for other fields of computer vision studies, the introduction of convolutional neural networks (CNNs) has dramatically advanced SR's performance. For instance, the pioneering work of the super-resolution convolution neural network (SRCNN) proposed by Dong~\etal~\cite{SRCNN,SRCNN-Ex} employed three convolutional layers to approximate the nonlinear mapping function from interpolated LR image to HR image and outperformed most conventional SR methods~\cite{PCPE-MDSISR_IS16,Hadamard_IS18}. Various works~\cite{FSRCNN,VDSR,DRCN,ESPCN,DRRN,LapSRN,EDSR,MemNet,SRDenseNet,IDN,RDN,RCAN} that explore network architecture designs and training strategies have continuously improved SR performance in terms of quantitative quality such as peak signal-to-noise ratio (PSNR), root mean squared error (RMSE), and structural similarity (SSIM)~\cite{SSIM}. However, these PSNR-oriented approaches still suffer from blurry results at large upscaling factors,~\eg, $4 \times$, particularly concerning the restoration of delicate texture details in the original HR image, distorted in the LR image.

In recent years, several perceptual-related methods have been exploited to boost visual quality under large upscaling factors~\cite{Perceptual-loss,SRGAN,EnhanceNet,CX,ESRGAN}. Specifically, the perceptual loss is proposed by Johnson~\etal~\cite{Perceptual-loss}, which is a loss function that measures differences of the intermediate features of VGG19~\cite{VGG19} when taking the ground-truth and generated images as inputs. Legig~\etal~\cite{SRGAN} extend this idea by adding an adversarial loss~\cite{GAN} and Sajjadi~\etal~\cite{EnhanceNet} combine perceptual, adversarial and texture synthesis losses to produce sharper images with realistic textures. Wang~\etal~\cite{SFTGAN} incorporate semantic segmentation maps into a CNN-based SR network to generate realistic and visually pleasing textures. Although these methods can produce sharper images, they typically contain artifacts that are readily observed. 

Moreover, these approaches tend to improve visual quality without considering the substantial degradation of quantitative quality. Since the primary objective of the super-resolution task is to make the enlarged images resemble the ground-truth HR images as much as possible, it is necessary to maintain nature while guaranteeing the basic structural features that is related to pixel-to-pixel losses~\eg, mean squared error (MSE), mean absolute error (MAE). At present, the most common way is to pre-train a PSNR-oriented model and then fine-tune this pre-trained model, in company with a discriminator network and perceptual loss. Even though this strategy helps increase the stability of the training process, it still requires updating all parameters of the generator, which means an increase in training time.

\begin{figure}[htpb]
	\begin{center}
		\includegraphics[width=0.6\textwidth]{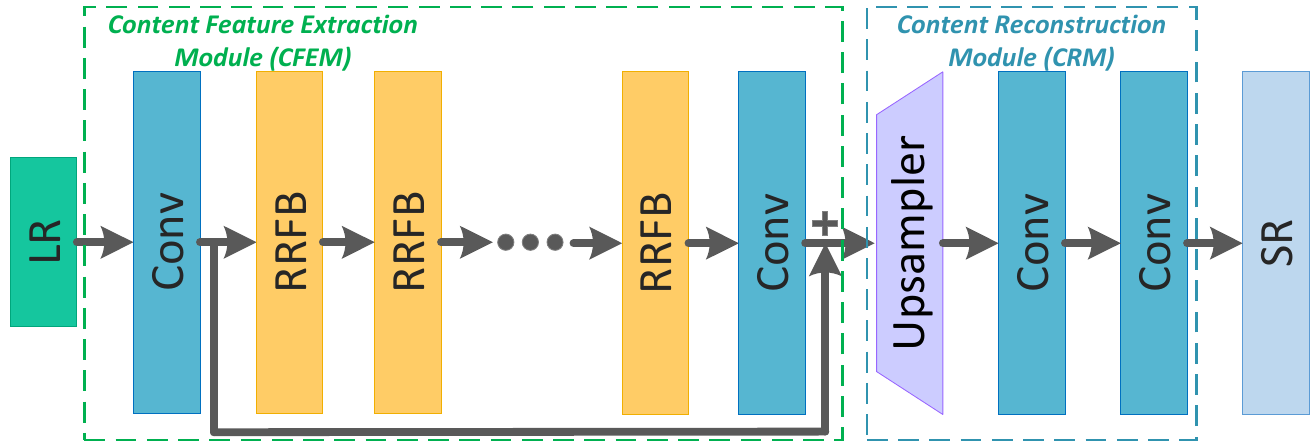}
	\end{center}
	\caption{The network architecture of our basic PSNR-oriented model (Residual Fusion Network, namely RFN). We use 24 RRFBs for our experiments.}
	\label{fig:basic_structure}
\end{figure}

\begin{figure}[htpb]
	\begin{center}
		\includegraphics[width=0.5\textwidth]{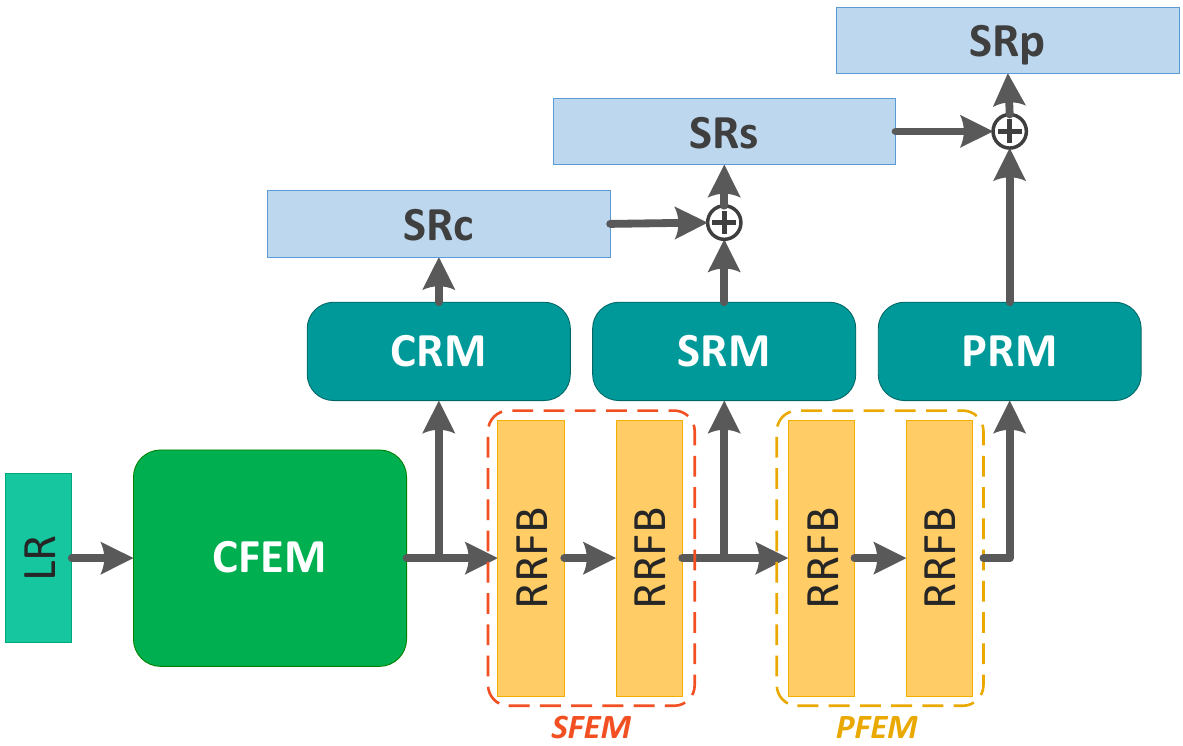}
	\end{center}
	\caption{The architecture of our progressive perception-oriented network (PPON). CFEM indicates content feature extraction module in Figure~\ref{fig:basic_structure}. CRM, SRM, and PRM represent content reconstruction module, structural reconstruction module, and photo-realism reconstruction module, respectively. SFEM denotes structural features extraction module and PFEM describes the perceptual features extraction part. In addition, $ \oplus $ is the element-wise summation operator.}
	\label{fig:architecture}
\end{figure}

\begin{figure*}[htpb]
	\begin{center}
		\subfigure[Hierarchical Feature Fusion Block (HFFB)]{\label{fig:HFFB}
			\includegraphics[width=0.45\textwidth]{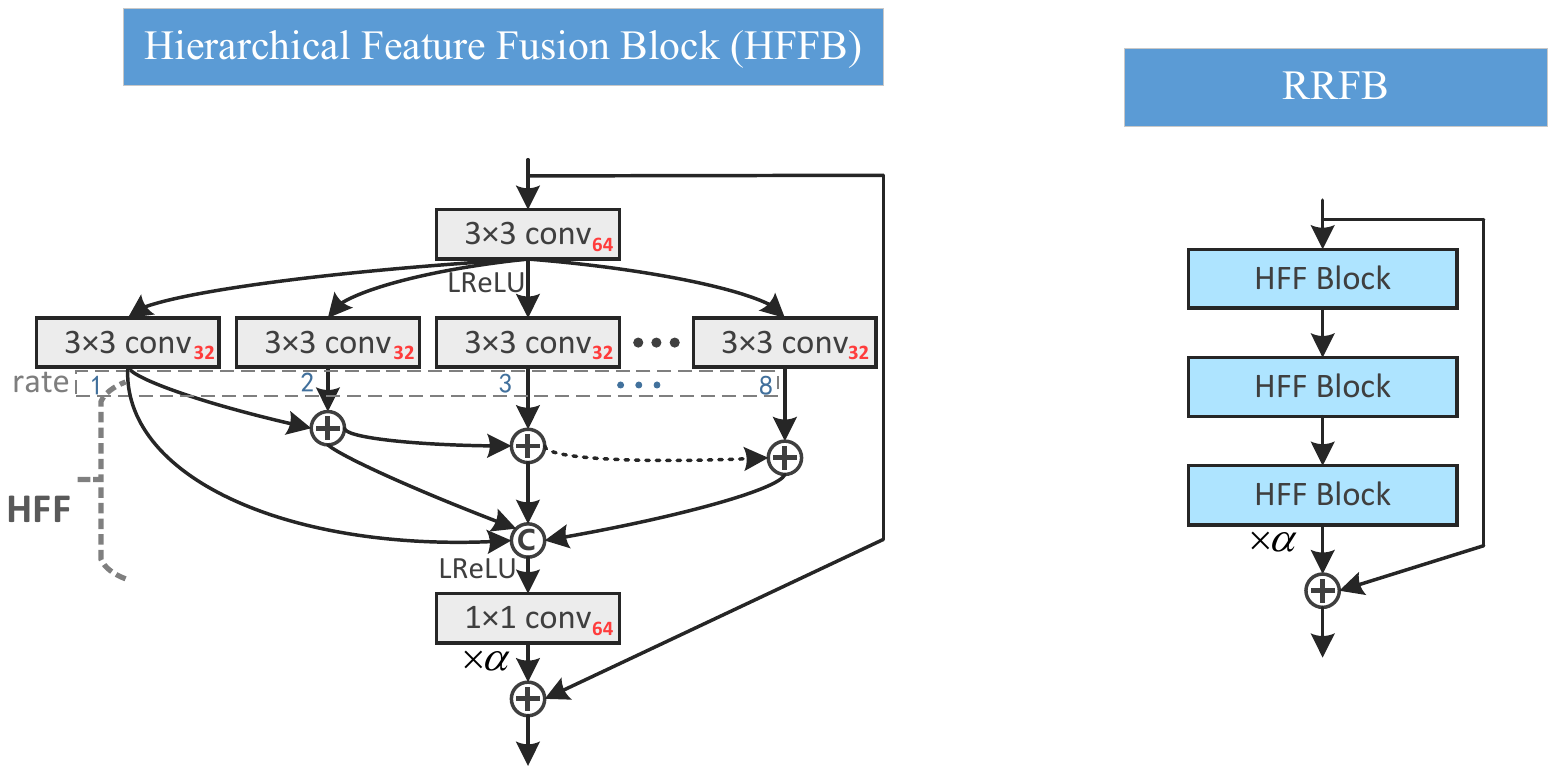}}
		\hfil
		\subfigure[Residual-in-Residual Fusion Block (RRFB)]{\label{fig:RRFB}
			\includegraphics[width=0.45\textwidth]{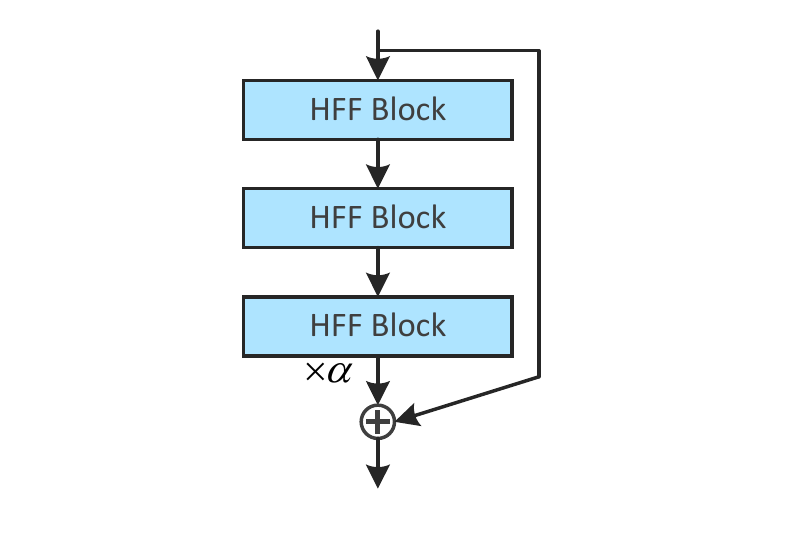}}
		\caption{The basic blocks are proposed in this work. (a) We employ 8 dilated convolutions. Each of them has 32 output channels for reducing block parameters. (b) RRFB is used in our primary and perception-oriented models and $\alpha$ is the residual scaling parameter~\cite{EDSR,ESRGAN}.}
		\label{fig:block}
	\end{center}
\end{figure*}

In this paper, we propose a novel super-resolution method via the progressive perception-oriented network (PPON), which gradually generates images with pleasing visual quality. More specifically, inspired by~\cite{ESPNet}, we propose a hierarchical feature fusion block (HFFB) as the basic block (shown in Figure~\ref{fig:HFFB}), which utilizes multiple dilated convolutions with different rates to exploit abundant multi-scale information. In order to ease the training of very deep networks, we assemble our basic blocks by using residual-in-residual fashion~\cite{RCAN,ESRGAN} named residual-in-residual fusion block (RRFB) as illustrated in Figure~\ref{fig:RRFB}. Our method adopts three reconstruction modules: a content reconstruction module (CRM), a structure reconstruction module (SRM), and a photo-realism reconstruction module (PRM). The CRM as showed in Figure~\ref{fig:basic_structure} mainly restores global information and minimizes pixel-by-pixel errors as previous PSNR-oriented approaches. The purpose of SRM is to maintain favorable structural information based on CRM's result using structural loss. Analogously, PRM estimates the residual between the real image and the output of SRM with adversarial and perceptual losses. The diagrammatic sketch of this procedure is given in Figure~\ref{fig:architecture}. Since the input of the perceptual features extraction module (PFEM) contains fruitful structure-related features and the generated perceptual image is built on the result of SRM, our PPON can synthesize a visually pleasing image that provides not only high-frequency components but also structural elements.

%% should be revised
To achieve rapid training, we develop a step-by-step training mode,~\ie, our basic model (illustrated in Figure~\ref{fig:basic_structure}) is trained first, then we freeze its parameters and train the sequential SFEM and SRM, and so on. The advantage is that when we train perception-related modules (PFEM and PRM), very few parameters need to be updated. It differs from previous algorithms that they require to optimize all parameters to produce photo-realistic results. Thus, it will reduce training time. 

Overall, our contributions can be summarized as follows.
\begin{itemize}
	
	\item We develop a progressive photo-realism reconstruction approach, which can synthesize images with high fidelity (PSNR) and compelling visual effects. Specifically, we develop three reconstruction modules for completing multiple tasks,~\ie, the content, structure, and perception reconstructions of an image. More broadly, we can also generate three images with different types in a feed-forward process, which is instructive to satisfy various task's requirements.
	
	\item We design an effective training strategy according to the characteristic of our proposed progressive perception-oriented network (PPON), which is to fix the parameters of the previous training phase and utilize the features produced by this trained model to update a few parameters at the current stage. In this way, the training of the perception-oriented model is robust and fast.
	
	\item We also propose the basic model RFN mostly constructed by cascading residual-in-residual fusion blocks (RRFBs), which achieves state-of-the-art performance in terms of PSNR.
	
\end{itemize}
The rest of this paper is organized as follows. Section~\ref{sec:related-work} provides a brief review of related SISR methods. Section~\ref{sec:proposed-method} describes the proposed approach and loss functions in detail. In Section~\ref{sec:experiments}, we explain the experiments conducted for this work, experimental comparisons with other state-of-the-art methods, and model analysis. In Section~\ref{sec:conclusion}, we conclude the study.

\section{Related Work}\label{sec:related-work}
In this section, we focus on deep neural network approaches to solve the SR problem.
\subsection{Deep learning-based super-resolution}
The pioneering work was done by Dong~\etal~\cite{SRCNN,SRCNN-Ex}, who proposed SRCNN for the SISR task, which outperformed conventional algorithms. To further improve the accuracy, Kim~\etal proposed two deep networks,~\ie, VDSR~\cite{VDSR}, and DRCN~\cite{DRCN}, which apply global residual learning and recursive layer respectively to the SR problem. Tai~\etal~\cite{DRRN} developed a deep recursive residual network (DRRN) to reduce the model size of the very deep network by using a parameter sharing mechanism. Another work designed by the authors is a very deep end-to-end persistent memory network (MemNet)~\cite{MemNet} for image restoration task, which tackles the long-term dependency problem in the previous CNN architectures. The methods mentioned above need to take the interpolated LR images as inputs. It inevitably increases the computational complexity and often results in visible reconstruction artifacts~\cite{LapSRN}.

To speed up the execution time of deep learning-based SR approaches, Shi~\etal~\cite{ESPCN} proposed an efficient sub-pixel convolutional neural network (ESPCN), which extracts features in the LR space and magnifies the spatial resolution at the end of the network by conducting an efficient sub-pixel convolution layer. Afterward, Dong~\etal~\cite{FSRCNN} developed a fast SRCNN (FSRCNN), which employs the transposed convolution to upscale and aggregate the LR space features. However, these two methods fail to learn complicated mapping due to the limitation of the model capacity. EDSR~\cite{EDSR}, the winner solution of NTIRE2017~\cite{NTIRE2017}, was presented by Lim~\etal. This work is much superior in performance to previous models. To alleviate the difficulty of SR tasks with large scaling factors such as $8 \times$, Lai~\etal~\cite{LapSRN} proposed the LapSRN, which progressively reconstructs the multiple SR images with different scales in one feed-forward network. Liu~\etal~\cite{MSDEPC_IS19} used the phase congruency edge map to guide an end-to-end multi-scale deep encoder and decoder network for SISR. Tong~\etal~\cite{SRDenseNet} presented a network for SR by employing dense skip connections, which demonstrated that the combination of features at different levels is helpful for improving SR performance. Recently, Zhang~\etal~\cite{RDN} extended this idea and proposed a residual dense network (RDN), where the kernel is residual dense block (RDB) that extracts abundant local features via dense connected convolutional layers. Furthermore, the authors proposed very deep residual channel attention networks (RCAN)~\cite{RCAN} that verified that the very deep network can availably improve SR performance and advantages of channel attention mechanisms. To leverage the execution speed and performance, IDN~\cite{IDN} and CARN~\cite{CARN} were proposed by Hui~\etal and Ahn~\etal, respectively. More concretely, Hui~\etal constructed a deep but compact network, which mainly exploited and fused different types of features. And Ahn~\etal designed a cascading network architecture. The main idea is to add multiple cascading connections from each intermediary layer to others. Such connections help this model performing SISR accurately and efficiently.

\subsection{Super-resolution considering naturalness}
SRGAN~\cite{SRGAN}, as a landmark work in perceptual-driven SR, was proposed by Ledig~\etal. This approach is the first attempt to apply GAN~\cite{GAN} framework to SR, where the generator is composed of residual blocks. To improve the naturalness of the images, perceptual and adversarial losses were utilized to train the model in SRGAN. Sajjadi~\etal~\cite{EnhanceNet} explored the local texture matching loss and further improved the visual quality of the composite images. Park~\etal~\cite{SRFeat} developed a GAN-based SISR method that produced realistic results by attaching an additional discriminator that works in the feature domain. Mechrez~\etal~\cite{CX} defined the Contextual loss that measured the similarity between the generated image and a target image by comparing the statistical distribution of the feature space. Wang~\etal~\cite{ESRGAN} enhanced SRGAN from three key aspects: network architecture, adversarial loss, and perceptual loss. A variant of Enhanced SRGAN (ESRGAN) won the first place in the PIRM2018-SR Challenge~\cite{PIRM-SR}.

\section{Proposed Method}\label{sec:proposed-method}
\subsection{The proposed PSNR-oriented SR model}
The single image super-resolution aims to estimate the SR image $I^\text{SR}$ from its LR counterpart ${I^\text{LR}}$. An overall structure of the proposed basic model (RFN) is shown in Figure~\ref{fig:basic_structure}. This network mainly consists of two parts: content feature extraction module (CFEM) and reconstruction part, where the first part extracts content features for conventional image SR task (pursuing high PSNR value), and the second part naturally reconstructs ${I^\text{SR}}$ through the front features related to the image content. The first procedure could be expressed by

\begin{equation}\label{eq:CFENet}
{F_c} = {H_\text{CFE}}\left( {{I^\text{LR}}} \right),
\end{equation}
where ${H_\text{CFE}\left(  \cdot  \right)}$ denotes content feature extractor,~\ie, CFEM. Then, ${F_c}$ is sent to the content reconstruction module (CRM) ${H_\text{CR}}$,
\begin{equation}\label{eq:CRNet}
{I_c^\text{SR}} = {H_\text{CR}}\left( {{F_c}} \right) = {H_\text{RFN}}\left( {{I^\text{LR}}} \right) ,
\end{equation}
where ${H_\text{RFN}\left( \cdot \right)}$ denotes the function of our RFN.

The basic model is optimized with the MAE loss function, followed by the previous works~\cite{EDSR,RDN,RCAN}. Given a training set $\left\{ {I_i^\text{LR},I_i^\text{HR}} \right\}_{i = 1}^N$, where $N$ is the number of training images, ${I_i^\text{HR}}$ is the ground-truth high-resolution image of the low-resolution image ${I_i^\text{LR}}$, the loss function of our basic SR model is

\begin{equation}\label{eq:l1-loss}
{{\cal L}_{content}}\left( \Theta_c  \right) = \frac{1}{N}\sum\limits_{i = 1}^N {{{\left\| {{H_\text{RFN}}\left( {I_i^\text{LR}} \right) - I_i^\text{HR}} \right\|}_1}} ,
\end{equation}
where $\Theta_c$ denotes the parameter set of our content-oriented branch (COBranch),~\ie, RFN.
\subsection{Progressive perception-oriented SR model}\label{branch}
\begin{figure}[htpb]
	\begin{center}
		\includegraphics[width=0.5\textwidth]{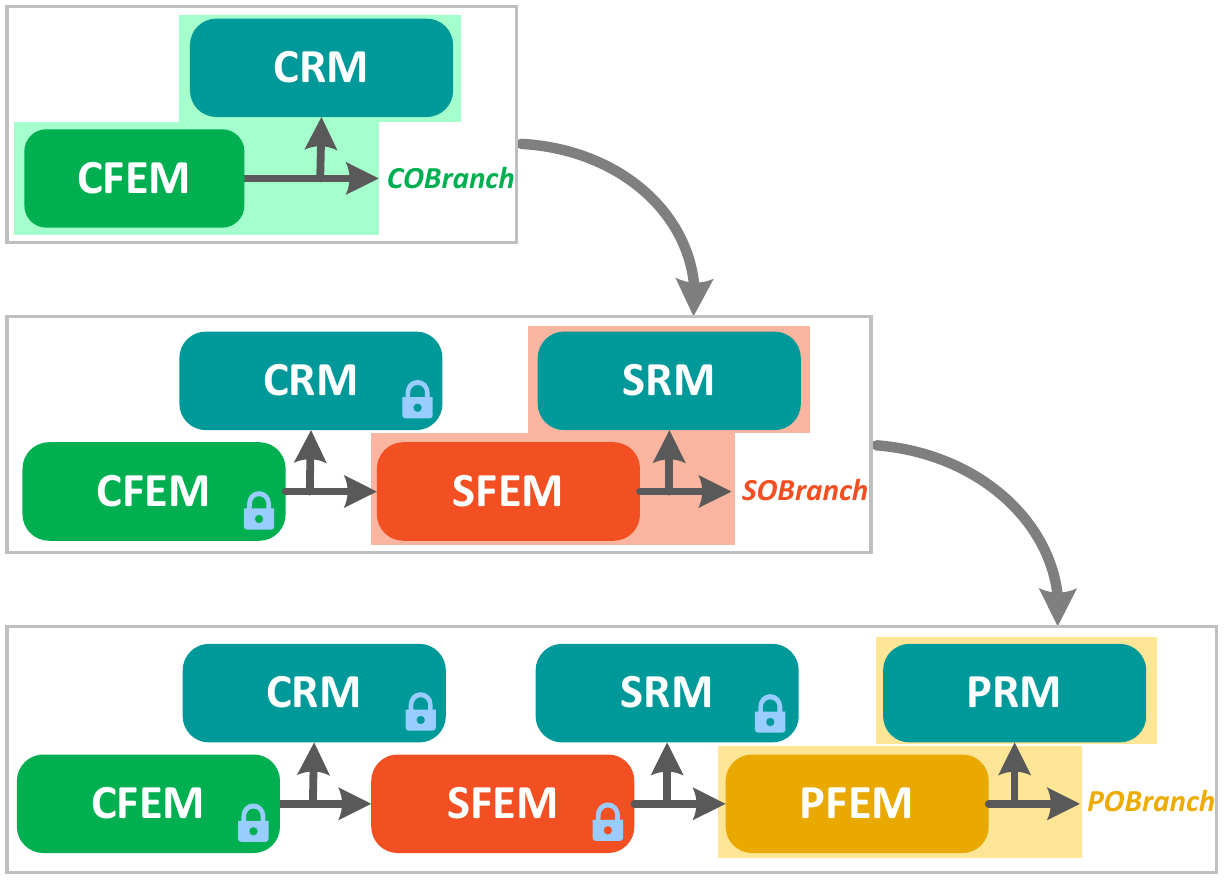}
	\end{center}
	\caption{The training scheme for our PPON. The light green region (COBranch) in the first row is our basic model RFN. Light red and yellow areas represent SOBranch and POBranch mentioned in Section~\ref{branch}, respectively. The entire training process is split into 3 stages. The module with miniature lock means to freeze its parameters.}
	\label{fig:training-scheme}
\end{figure}

As depicted in Figure~\ref{fig:architecture}, based on the content features extracted by the CFEM, we design a SFEM to distill structure-related information for restoring images with SRM. This process can be expressed by

\begin{equation}\label{eq:bo-branch}
I_s^\text{SR} = {H_\text{SR}}\left( {{F_s}} \right) + I_c^\text{SR} = {H_\text{SR}}\left( {{H_\text{SFE}}\left( {{F_c}} \right)} \right) + I_c^\text{SR} ,
\end{equation}
where ${H_\text{SR}}\left(  \cdot  \right)$ and ${{H_\text{SFE}}\left( \cdot \right)}$ denote the functions of SRM and SFEM, respectively. To this end, we employ the multi-scale structural similarity index (MS-SSIM) and multi-scale ${L_1}$ as loss functions to optimize this branch. SSIM is defined as

\begin{equation}\label{eq:SSIM}
SSIM\left( {x,y} \right) = \frac{{2{\mu _x}{\mu _y} + {C_1}}}{{\mu _x^2 + \mu _y^2 + {C_1}}} \cdot \frac{{2{\sigma _{xy}} + {C_2}}}{{\sigma _x^2 + \sigma _y^2 + {C_2}}} = l\left( {x,y} \right) \cdot cs\left( {x,y} \right),
\end{equation}
where ${\mu _x}$, ${\mu _y}$ are the mean, ${{\sigma _{xy}}}$ is the covariance of $x$ and $y$, and ${C_1}$, ${C_2}$ are constants. Given multiple scales through a process of $M$ stages of downsampling, MS-SSIM is defined as
\begin{equation}\label{eq:MS-SSIM}
\text{MS-SSIM}(x,y) = l_M^\alpha \left( {x,y} \right) \cdot \prod\limits_{j = 1}^M {cs_j^{{\beta _j}}\left( {x,y} \right)} ,
\end{equation}
where ${l_M}$ and ${cs}_j$ are the term we defined in Equation~\ref{eq:SSIM} at scale $M$ and $j$, respectively. From~\cite{MS-SSIM}, we set $\alpha  = {\beta _M}$ and $\beta  = \left[ {0.0448,0.2856,0.3001,0.2363,0.1333} \right]$. Therefore, the total loss function of our structure branch can be expressed by

\begin{equation}\label{eq:MS-SSIM-loss}
{{\cal L}_\text{MS-SSIM}} = \frac{1}{N}\sum\limits_{i = 1}^N {\left[ {1 - \text{MS-SSIM}\left( {I_i^{HR},{H_\text{SOB}}\left( {F_c^i} \right)} \right)} \right]} ,
\end{equation}
where ${H_\text{SOB}}\left(  \cdot  \right)$ represents the cascade of SFEM and SRM (light red area in Figure~\ref{fig:training-scheme}). $F_c^i$ denotes content features (see Equation~\ref{eq:CFENet}) corresponding to $i$-th training sample in a batch. Thus, the total loss function of this branch can be formulated as follows

\begin{equation}\label{eq:structure-loss}
{{\cal L}_{structure}\left(\Theta_s \right)} = {{\cal L}_\text{MS-L1}} + \lambda {{\cal L}_\text{MS-SSIM}} ,	
\end{equation}
where ${{\cal L}_\text{MS-L1}} = \sum\limits_{j = 1}^M {{\omega _j} \cdot {l_{mae}}\left( {{x_j},{y_j}} \right)} $ and $ \lambda$ is a scalar value to balance two losses, $\Theta_s$ denotes the parameter set of structure-oriented branch (SOBranch). Here, we set $M = 5$, ${\omega_{1,2,...,5}} = [1, 0.5, 0.25, 0.125, 0.125]$ through experience.

Similarly, to obtain photorealistic images, we utilize structural-related features refined by SFEM and send them to our perception feature extraction module (PFEM). The merit of this practice is to avoid re-extracting features from the image domain. These extracted features contain abundant and superior quality structural information, which tremendously helps perceptual-oriented branch (POBranch, see in Figure~\ref{fig:training-scheme}) generate visually plausible SR images while maintaining the basic structure. Concretely, structural feature $F_s$ is entered in PFEM

\begin{equation}\label{eq:po-branch}
I_p^\text{SR} = {H_\text{PR}}\left( {{F_p}} \right) + I_s^\text{SR} = {H_\text{PR}}\left( {{H_\text{PFE}}\left( {{F_s}} \right)} \right) + I_s^\text{SR} ,
\end{equation}
where ${H_\text{PR}}\left( \cdot \right)$ and ${{H_\text{PFE}}\left( \cdot \right)}$ indicate PRM and PFEM as shown in Figure~\ref{fig:architecture}, respectively. For pursuing better visual effect, we adopt Relativistic GAN~\cite{RaSGAN} as in~\cite{ESRGAN}. Given a real image ${x_r}$ and a fake one ${x_f}$, the relativistic discriminator intends to estimate the probability that ${x_r}$ is more realistic than ${x_f}$. In standard GAN, the discriminator can be defined, in term of the non-transformed layer ${C\left( x \right)}$, as $D\left( x \right) = \sigma \left( {C\left( x \right)} \right)$, where $\sigma$ is sigmoid function. The Relativistic average Discriminator (RaD, denoted by $D_\text{Ra}$)~\cite{RaSGAN} can be formulated as ${D_\text{Ra}}\left( {{x_r},{x_f}} \right) = \sigma \left( {C\left( x \right) - {\mathbb{E}_{{x_f}}}\left[ {C\left( {{x_f}} \right)} \right]} \right)$, if $x$ is real. Here, ${{\mathbb{E}_{{x_f}}}\left[ {C\left( \cdot \right)} \right]}$ is the average of all fake data in a batch. The discriminator loss is defined by

\begin{equation}
{\cal L}_D^{Ra} =  -\mathbb{E}{_{{x_r}}}\left[ {\log \left( {{D_{Ra}}\left( {{x_r},{x_f}} \right)} \right)} \right] -\mathbb{E}{_{{x_f}}}\left[ {\log \left( {1 - {D_{Ra}}\left( {{x_f},{x_r}} \right)} \right)} \right].
\end{equation}
The corresponding adversarial loss for generator is
\begin{equation}
{\cal L}_G^{Ra} =  -\mathbb{E}{_{{x_r}}}\left[ {\log \left( {1 - {D_{Ra}}\left( {{x_r},{x_f}} \right)} \right)} \right] - \mathbb{E}{_{{x_f}}}\left[ {\log \left( {{D_{Ra}}\left( {{x_f},{x_r}} \right)} \right)} \right].
\end{equation}
where $x_f$ represents the generated images at the current perception-maximization stage,~\ie, $I_p^\text{SR}$ in equation~\ref{eq:po-branch}.

VGG loss that has been investigated in recent SR works~\cite{Perceptual-loss,SRGAN,EnhanceNet,ESRGAN} for better visual quality is also introduced in this stage. We calculate the VGG loss based on the ``conv5\_4'' layer of VGG19~\cite{VGG19},
\begin{equation}
{{\cal L}_{vgg}} = \frac{1}{V}\sum\limits_{i = 1}^C {{{\left\| {{\phi _i}\left( {{I^{HR}}} \right) - {\phi _i}\left( {I_p^{SR}} \right)} \right\|}_1}} ,
\end{equation}
where $V$ and $C$ indicate the tensor volume and channel number of the feature maps, respectively, and ${{\phi _i}}$ denotes the $i$-th channel of the feature maps extracted from the hidden layer of VGG19 model. Therefore, the total loss for the perception stage is:
\begin{equation}\label{eq:perception-loss}
{{\cal L}_{perception} \left( \Theta_p \right)} = {{\cal L}_{vgg}} + \eta {\cal L}_G^{Ra} ,
\end{equation}
where $\eta$ is the coefficients to balance these loss functions. And $\Theta_p$ is the training parameters of POBranch.
\subsection{Residual-in-residual fusion block}
\begin{figure}[htpb]
	\begin{center}
		\includegraphics[width=0.6\textwidth]{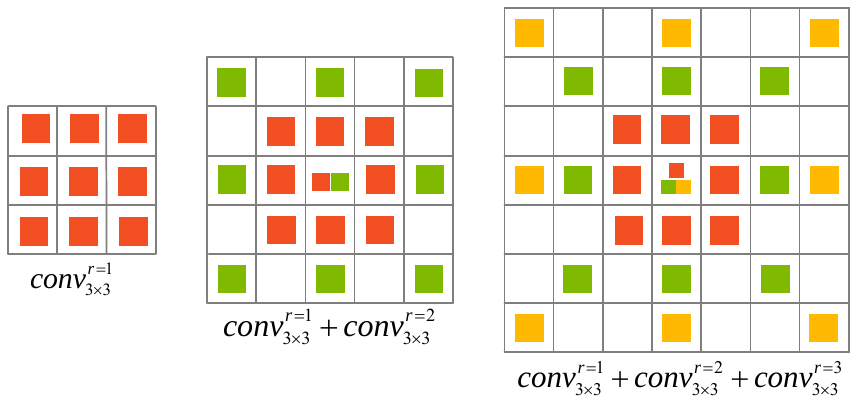}
	\end{center}
	\caption{The diagrammatic sketch of multiple dilated convolutions addition. Taking the middle sub-figure as an example, $conv_{3 \times 3}^{r = 2}$ indicates $3 \times 3$ dilated convolution with dilation rate of 2. Under the same conditions of receptive field, $conv_{3 \times 3}^{r = 1}+conv_{3 \times 3}^{r = 2}$ is more dense than $conv_{3 \times 3}^{r = 2}$.}
	\label{fig:sketch}
\end{figure}

We now give more details about our proposed RRFB structure (see Figure~\ref{fig:RRFB}), which consists of multiple hierarchical feature fusion blocks (HFFB) (see Figure~\ref{fig:RRFB}). Unlike the frequently-used residual block in SR, we intensify its representational ability by introducing the spatial pyramid of dilated convolutions~\cite{ESPNet}. Specifically, we develop $K$ $n \times n$ dilated convolutional kernels simultaneously, each with a dilation rate of $k$, $k = \left\{ {1, \ldots ,K} \right\}$. Due to these dilated convolutions preserve different receptive fields, we can aggregate them to obtain multi-scale features. As shown in Figure~\ref{fig:sketch}, single dilated convolution with a dilation rate of 3 (yellow block) looks sparse. The feature maps obtained using kernels of different dilation rates are hierarchically added to acquire an effective receptive field before concatenating them. A simple example is illustrated in Figure~\ref{fig:sketch}. For explaining this hierarchical feature fusion process clearly, the output of dilated convolution with a dilation rate of $k$ is denoted by $f_k$. In this way, concatenated multi-scale features $H_{ms}$ can be expressed by
\begin{equation}
{H_{ms}} = \left[ {{f_1},{f_1} + {f_2}, \ldots ,{f_1} + {f_2} +  \cdots  + {f_K}} \right] .
\end{equation}
After collecting these multi-scale features, we fuse them through a $1 \times 1$ convolution $Con{v_{1 \times 1}}$, that is $Con{v_{1 \times 1}}\left( {L{\mathop{\rm Re}\nolimits} LU\left( {{F_{ms}}} \right)} \right)$. Finally, the local skip connection with residual scaling is utilized to complete our HFFB.
\section{Experiments}\label{sec:experiments}
\subsection{Datasets and Training Details}
We use the DIV2K dataset~\cite{NTIRE2017}, which consists of 1,000 high-quality RGB images (800 training images, 100 validation images, and 100 test images) with 2K resolution. For increasing the diversity of training images, we also use the Flickr2K dataset~\cite{EDSR} consisting of 2,650 2K resolution images. In this way, we have 3,450 high-resolution images for training purposes. LR training images are obtained by downscaling HR with a scaling factor of $4 \times$ images using bicubic interpolation function in MATLAB. HR image patches with a size of $192 \times 192$ are randomly cropped from HR images as the input of our proposed model, and the mini-batch size is set to 25. Data augmentation is performed on the 3,450 training images, which are randomly horizontal flip and 90-degree rotation. For evaluation, we use six widely used benchmark datasets: Set5~\cite{Set5}, Set14~\cite{Set14}, BSD100~\cite{BSD100}, Urban100~\cite{Urban100}, Manga109~\cite{Manga109}, and the PIRM dataset~\cite{PIRM-SR}. The SR results are evaluated with PSNR, SSIM~\cite{SSIM}, learned perceptual image patch similarity (LPIPS)~\cite{LPIPS}, and perceptual index (PI) on Y (luminance) channel, in which PI is based on the non-reference image quality measures of Ma~\etal~\cite{Ma_Score} and NIQE~\cite{NIQE},~\ie, $\text{PI} = \frac{1}{2}\left( {\left( {10 - \text{Ma}} \right) + \text{NIQE}} \right)$. The lower values of LPIPS and PI, the better.

As depicted in Figure~\ref{fig:training-scheme}, the training process is composed of three phases. First, we train the COBranch with Equation~\ref{eq:l1-loss}. The initial learning rate is set to $2 \times {10^{ - 4}}$, which is decreased by 2 for every 1000 epochs ($1.38 \times {10^5}$ iterations). And then, we fix the parameters of COBranch and only train the SOBranch through the loss function in Equation~\ref{eq:structure-loss} with $\lambda  = 1 \times {10^3}$. This process is illustrated in the second row of Figure~\ref{fig:training-scheme}. During this stage, the learning rate is set to $1 \times {10^{ - 4}}$ and halved at every 250 epochs ($3.45 \times {10^4}$ iterations). Similarly, we eventually only train the POBranch by Equation~\ref{eq:perception-loss} with $\eta  = 5 \times {10^{ - 3}}$. The learning rate scheme is the same as the second phase. All the stages are trained by ADAM optimizer~\cite{Adam} with the momentum parameter ${\beta _1} = 0.9$. We apply the PyTorch v1.1 framework to implement our model and train them using NVIDIA TITAN Xp GPUs.

We set the dilated convolutions number as $K=8$ in the HFFB structure. All dilated convolutions have $3 \times 3$ kernels and 32 filters, as shown in Figure~\ref{fig:HFFB}. In each RRFB, we set the HFFB number as 3. In COBranch, we apply 24 RRFBs. Moreover, only 2 RRFBs are employed in both SOBranch and POBranch. All standard convolutional layers have 64 filters, and their kernel sizes are set to $3 \times 3$ expect for that at the end of HFFB, whose kernel size is $1 \times 1$.  The residual scaling parameter $\alpha=0.2$ and the negative slope of LReLU is set as $0.2$.

\subsection{Model analysis}
\begin{figure}[htbp]
	\begin{center}
		\includegraphics[width=0.8\textwidth]{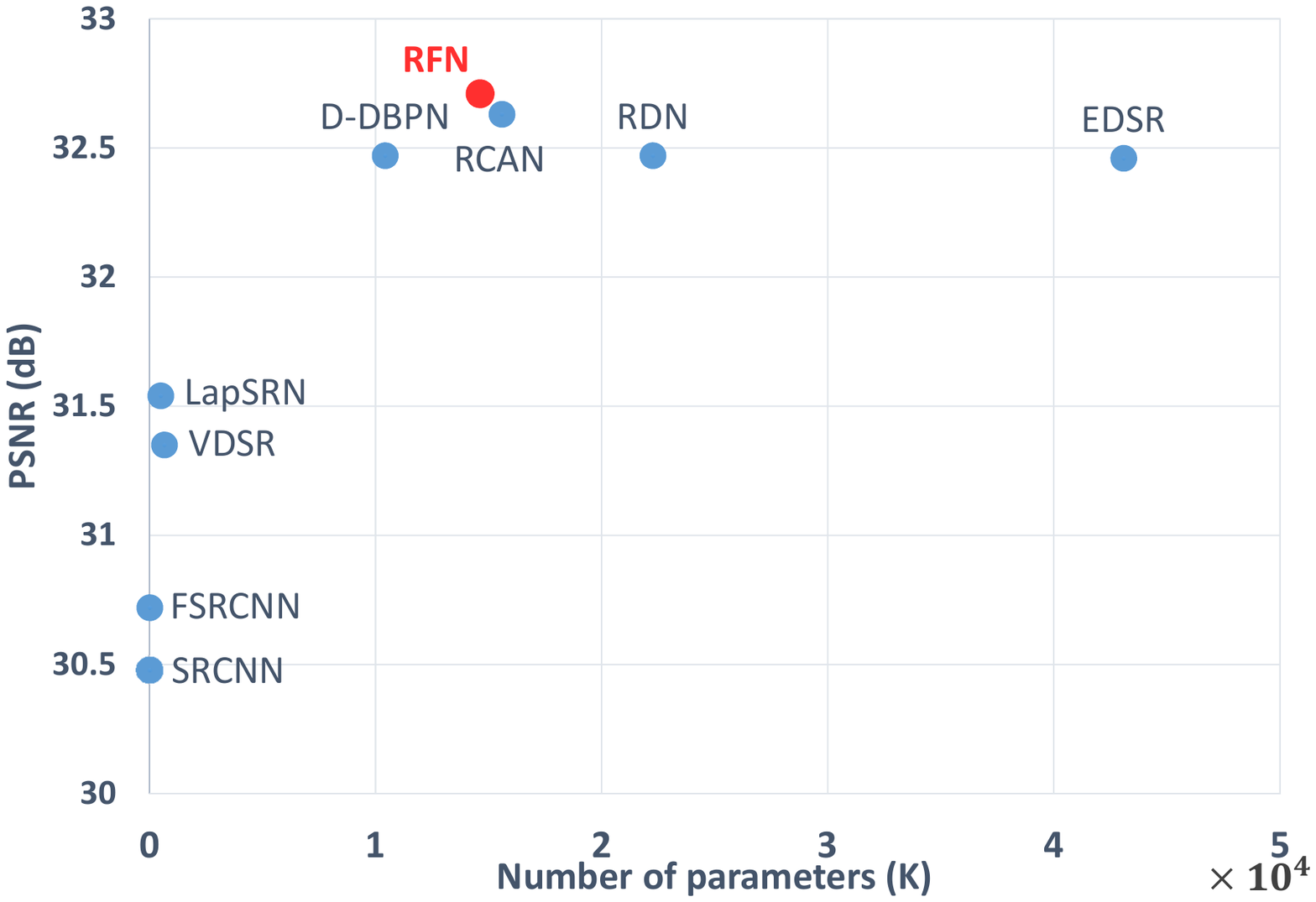}
	\end{center}
	\caption{PSNR performance and number of parameters. The results are evaluated on Set5 dataset for a scaling factor of $4 \times$.}
	\label{fig:parameters}
\end{figure}
\textbf{Model Parameters.} We compare the trade-off between performance and model size in Figure~\ref{fig:parameters}. Among the nine models, RFN and RCAN show higher PSNR values than others. In particular, RFN scores the best performance in Set5. It should be pointed out that RFN uses fewer parameters than RCAN to achieve this performance. It does mean that RFN can better balance performance and model size.

\textbf{Study of dilation convolution and hierarchical feature fusion.} We remove the hierarchical feature fusion structure. Furthermore, in order to investigate the function of dilated convolution, we use ordinary convolutions. For validating quickly, only $1$ RRFB is used in CFEM, and this network is called RFN\_mini. We conduct the training process with the DIV2K dataset, and the results are depicted in Table~\ref{tab:ablation-study}. As the number of RRFB increases, the benefits will increase accumulatively (see in Table~\ref{tab:ablation-study-multi}).

\begin{table}[htbp]
	\caption{Investigations of dilated convolution and hierarchical fusion. These models are trained 200k iterations with DIV2K training dataset.}
	\label{tab:ablation-study}
	\begin{center}
		\begin{tabular}{|c|c|c|c|c|}
			\hline
			Dilated convolution & \XSolidBrush & \XSolidBrush & \Checkmark & \Checkmark \\
			\hline
			Hierarchical fusion & \XSolidBrush & \Checkmark & \XSolidBrush & \Checkmark \\
			\hline
			\hline
			PSNR on Set5 $\left( 4 \times \right)$ & 31.68 & 31.69 & 31.63 & 31.72 \\
			\hline
		\end{tabular}
		
	\end{center}
\end{table}

%-----------------------------------------------------------------
% The effectiveness of dilated convolution and hierarchical fusion
%-----------------------------------------------------------------
\begin{table}[htb]
	\small
	\centering
	\caption{Investigations of dilated convolution. Above models are trained 300k iterations with DIV2K training dataset.}
	\label{tab:ablation-study-multi}
		\begin{tabular}{|c|c|c|c|c|c|}
			\hline
			Method & N\_blocks & Set5 & Set14 & BSD100 & Urban100 \\
			\hline
			w/o dilation & 2 & 32.05 & 28.51 & 27.52 & 25.91 \\
			RFN\_Mini & 2 & 32.07 & 28.53 & 27.53 & 25.91 \\
			\hline
			w/o dilation & 4 & 32.18 & 28.63 & 27.59 & 26.16 \\
			RFN\_Mini & 4 & 32.26 & 28.67 & 27.60 & 26.23 \\
			\hline 
	\end{tabular}
\end{table}

\begin{figure*}[htpb]
	\centering
	\scalebox{0.9}{
		\begin{tabular}{cc}
			\begin{adjustbox}{valign=t}
				\begin{tabular}{c}
					\includegraphics[width=0.2\textwidth, height=0.15\textheight]{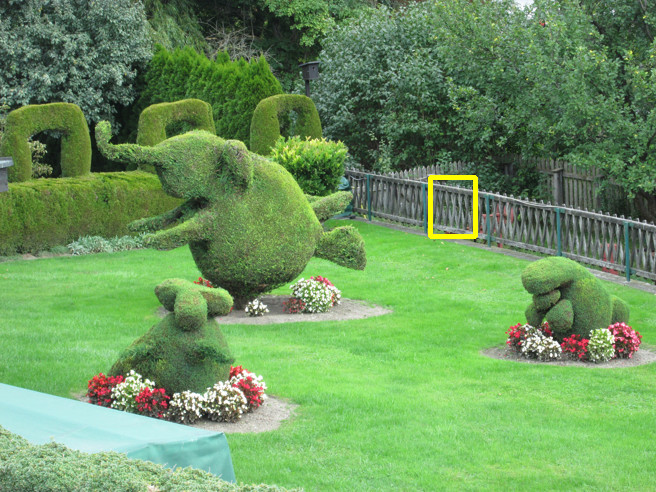} \\
					PIRM\_Val: 71 \\
				\end{tabular}
			\end{adjustbox}
			\hspace{-4mm}
			\begin{adjustbox}{valign=t}
				\begin{tabular}{cccc}
					\includegraphics[width=0.2\textwidth, height=0.15\textheight]{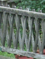} & 
					\hspace{-4mm}
					\includegraphics[width=0.2\textwidth, height=0.15\textheight]{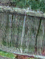} &
					\hspace{-4mm}
					\includegraphics[width=0.2\textwidth, height=0.15\textheight]{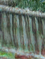} &
					\hspace{-4mm}
					\includegraphics[width=0.2\textwidth, height=0.15\textheight]{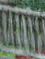} \\
					HR &\hspace{-4mm} (a) &\hspace{-4mm} (b) &\hspace{-4mm}  PPON \\
				\end{tabular}
			\end{adjustbox}
	\end{tabular} }
	\caption{Ablation study of progressive structure. (a) w/o CRM \& SOBranch. (b) w/o SOBranch.}
	\label{fig:ablation-study-progressive-structure}
\end{figure*}

\begin{figure*}[ht]
	\begin{center}
		\includegraphics[width=0.98\textwidth]{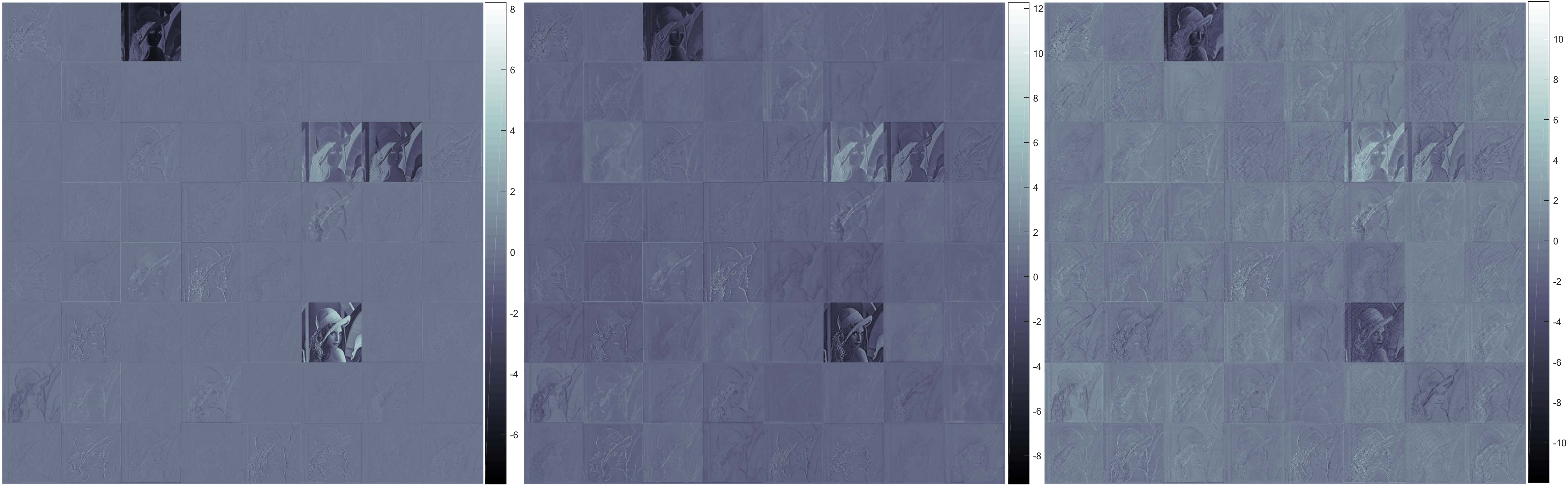}
	\end{center}
	\caption{The feature maps of CFEM, SFEM, and PFEM are visualized from left to right. Best viewed with zoom-in.}
	\label{fig:visualization}
\end{figure*}

\begin{table*}[htpb]
	\centering
	\caption{Ablation study of progressive structure (with GAN). PSNR, SSIM, and PI are evaluated on the Y channel while LPIPS are conducted on the RGB color space.}
	\label{tab:ablation-study-structure}
	\scalebox{0.5}{
	\begin{tabular}{|c|c|c|c|}
		\hline
		Item & w/o CRM \& SOBranch & w/o SOBranch & PPON \\
		\hline
		\hline
		Memory footprint (M) & 11,599 & 5,373 & 5,357 \\
		Training time (sec/epoch) & 347 & 176 & 183\\
		PIRM\_Val (PSNR / SSIM / LPIPS / PI) & 25.61 / 0.6802 / 0.1287 / 2.2857 & \textbf{26.32} / 0.6981 / 0.1250 / \textbf{2.2282} & 26.20 / \textbf{0.6995} / \textbf{0.1194} / 2.2353 \\
		PIRM\_Test (PSNR / SSIM/ LPIPS / PI) & 25.47 / 0.6667 / 0.1367 / 2.2055 & \textbf{26.16} / 0.6831 / 0.1309 / 2.1704 & 26.01 / \textbf{0.6831} / \textbf{0.1273} / \textbf{2.1511} \\
		\hline
	\end{tabular} }
\end{table*}

\begin{table*}[htpb]
	\centering
	\caption{Performance of RFN and S-RFN (without GAN). All metrics are performed on the RGB color space.}
	\label{tab:performance}
	\scalebox{0.85}{
	\begin{tabular}{|l|c|c|}
		\hline
		Item & RFN & S-RFN \\
		\hline
		\hline
		Memory footprint (M) & 8,799 & 2,733 \\
		Training time (sec/epoch) & 278 & 110 \\
		PIRM\_Val (PSNR / SSIM / LPIPS) & \textbf{27.27} / 0.8961 / 0.2901 & 27.14 / \textbf{0.7741} / \textbf{0.2651} \\
		PIRM\_Test (PSNR / SSIM / LPIPS) & \textbf{27.14} / 0.7571 / 0.3077 & 27.00 / \textbf{0.7637} / \textbf{0.2804} \\
		Set5 (PSNR / SSIM / LPIPS) & \textbf{30.68} / 0.8714 / 0.1709 & 30.62 / \textbf{0.8737} / \textbf{0.1684} \\
		Set14 (PSNR / SSIM / LPIPS) & \textbf{26.88} / 0.7543 / 0.2748 & 26.76 / \textbf{0.7595} / \textbf{0.2583} \\
		B100 (PSNR / SSIM / LPIPS) & \textbf{26.52} / 0.7225 / 0.3620 & 26.40 / \textbf{0.7302} / \textbf{0.3377} \\
		Urban100 (PSNR / SSIM / LPIPS) & \textbf{25.46} / 0.7940 / 0.1982 & 25.39 / \textbf{0.7982} / \textbf{0.1879} \\
		Manga109 (PSNR / SSIM / LPIPS) & \textbf{29.71} / 0.8945 / 0.0984 & 29.62 / \textbf{0.8961} / \textbf{0.0939} \\
		\hline
	\end{tabular}
	}
\end{table*}

\subsection{Progressive structure analysis}
\begin{figure*}[ht]
	\centering
	\scriptsize
	\scalebox{0.86}{
	\begin{tabular}{ccccccc}
		HR & SRGAN~\cite{SRGAN} & ENet~\cite{EnhanceNet} & CX~\cite{CX} & SuperSR~\cite{ESRGAN} & ESRGAN~\cite{ESRGAN} & PPON(Ours) \\
		\includegraphics[width=0.15\textwidth,height=0.13\textwidth]{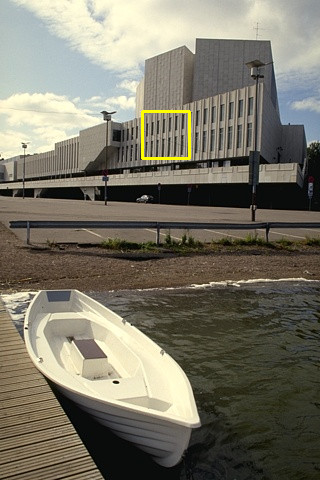} &
		\hspace{-3mm}
		\includegraphics[width=0.15\textwidth,height=0.13\textwidth]{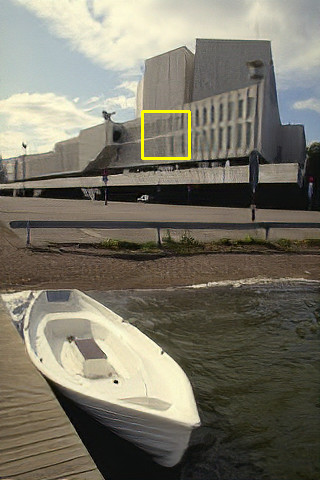} &
		\hspace{-3mm}
		\includegraphics[width=0.15\textwidth,height=0.13\textwidth]{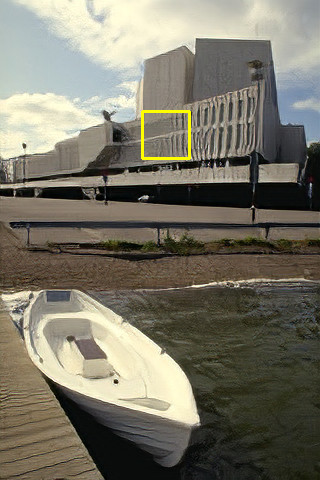} & 
		\hspace{-3mm}
		\includegraphics[width=0.15\textwidth,height=0.13\textwidth]{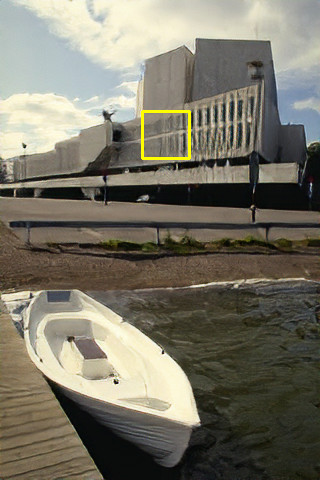} & 
		\hspace{-3mm}
		\includegraphics[width=0.15\textwidth,height=0.13\textwidth]{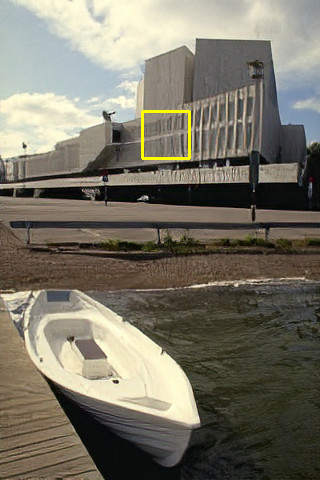} & 
		\hspace{-3mm}
		\includegraphics[width=0.15\textwidth,height=0.13\textwidth]{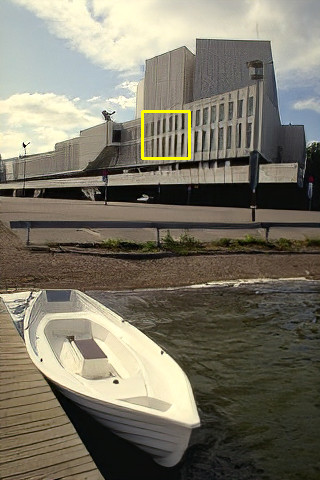} & 
		\hspace{-3mm}
		\includegraphics[width=0.15\textwidth,height=0.13\textwidth]{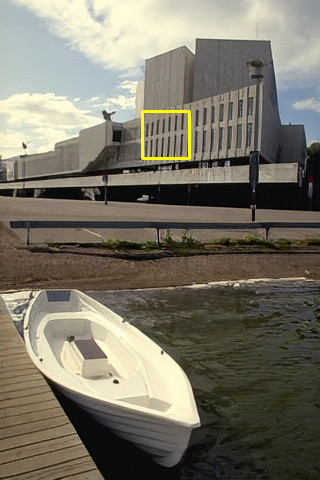} \\
		
		\includegraphics[width=0.15\textwidth,height=0.13\textwidth]{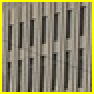} &
		\hspace{-3mm}
		\includegraphics[width=0.15\textwidth,height=0.13\textwidth]{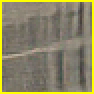} &
		\hspace{-3mm}
		\includegraphics[width=0.15\textwidth,height=0.13\textwidth]{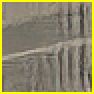} &
		\hspace{-3mm}
		\includegraphics[width=0.15\textwidth,height=0.13\textwidth]{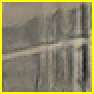} &
		\hspace{-3mm}
		\includegraphics[width=0.15\textwidth,height=0.13\textwidth]{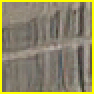} &
		\hspace{-3mm}
		\includegraphics[width=0.15\textwidth,height=0.13\textwidth]{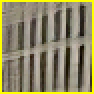} &
		\hspace{-3mm}
		\includegraphics[width=0.15\textwidth,height=0.13\textwidth]{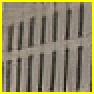} \\
	\end{tabular} }
	\caption{An example of the structure distortion. The image is from the BSD100 dataset~\cite{BSD100}.}
	\label{fig:distortion-example}
\end{figure*}

\begin{figure*}[ht]
	\centering
	\begin{tabular}{ccc}
		SRc & SRs & SRp \\
		\includegraphics[width=0.31\textwidth]{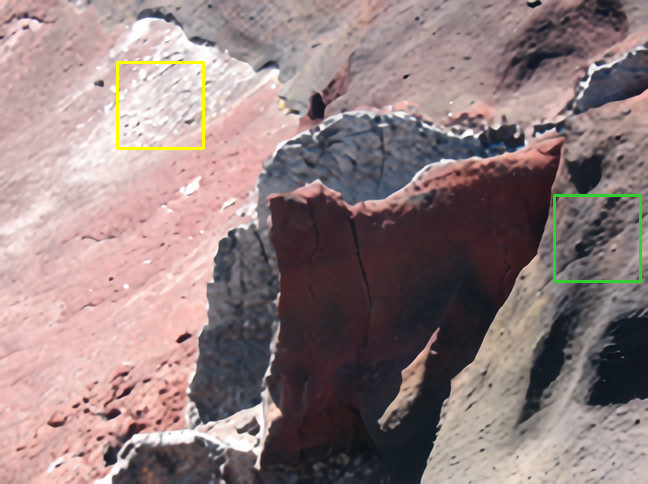} &
		\includegraphics[width=0.31\textwidth]{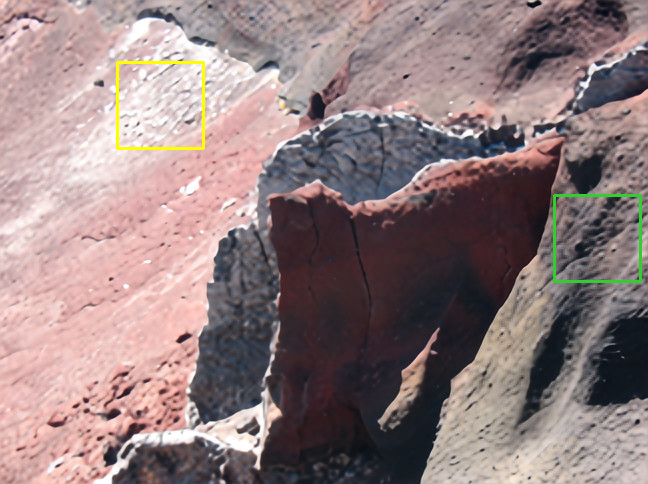} &
		\includegraphics[width=0.31\textwidth]{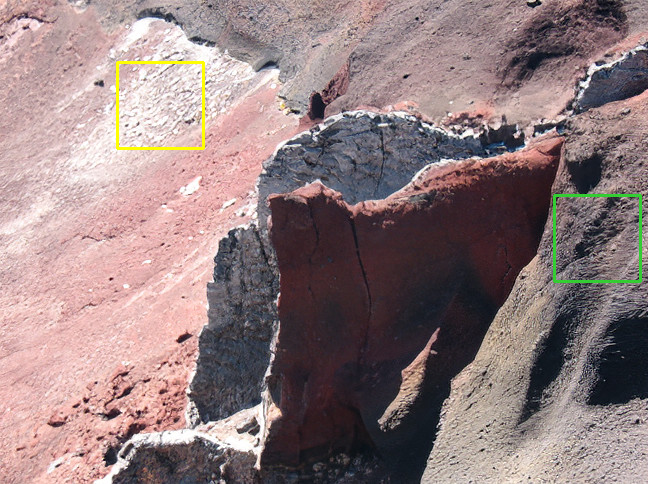} \\
	\end{tabular}
	\begin{tabular}{ccc}
		\includegraphics[width=0.149\textwidth]{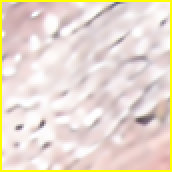} 
		\includegraphics[width=0.149\textwidth]{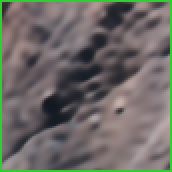} &
		\includegraphics[width=0.15\textwidth]{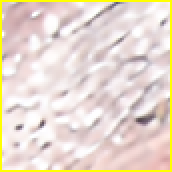} 
		\includegraphics[width=0.15\textwidth]{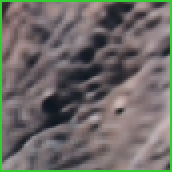} &
		\includegraphics[width=0.151\textwidth]{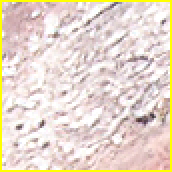} 
		\includegraphics[width=0.151\textwidth]{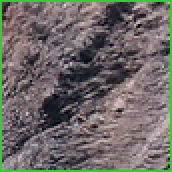} \\
	\end{tabular}
	\caption{A comparison of the visual effects between the three branch outputs. SRc, SRs, and SRp are outputs of the COBranch, SOBranch, and POBranch, respectively. The image is from the PIRM\_Val dataset~\cite{PIRM-SR}.}
	\label{fig:example}
\end{figure*}

We observe that perceptual-driven SR results produced by GAN-based approaches~\cite{SRGAN,EnhanceNet,CX} often suffer from structural distortion, as illustrated in Figure~\ref{fig:distortion-example}. To alleviate this problem, we explicitly add structural information through our devised progressive architecture described in the main manuscript. To make it easier to understand this progressive practice, we show an example in Figure~\ref{fig:example}. From this picture, we can note that the difference between SRc and SRp is mainly reflected in the sharper texture of SRp. Therefore, the remaining component is substantially the same. Please take into account this viewpoint, we naturally design the progressive topology structure,~\ie, gradually adding high-frequency details. 

To validate the feature maps extracted by the CFEM, SFEM, and PFEM have dependencies and relationships, we visualize the intermediate feature maps, as shown in Figure~\ref{fig:visualization}. From this picture, we can find that the feature maps distilled by three different extraction modules are similar. Thus, features extracted in the previous stage can be utilized in the current phase. In addition, feature maps in the third sub-figure contain more texture information, which is instructive to the reconstruction of visually high-quality images. To verify the necessity of using progressive structure, we remove CRM and SOBranch from PPON (\ie, changing to normal structure, similar to ESRGAN~\cite{ESRGAN}). We observe that PPON without CRM \& SOBranch cannot generate clear structural information, while PPON can better recover it. Table~\ref{tab:ablation-study-structure} suggests that our progressive structure can significantly improve the fidelity measured by PSNR and SSIM while improving perceptual quality. It indicates that fewer updatable parameters not only occupy less memory but also encourage faster training. ``w/o CRM \& SOBranch'' is a fundamental architecture without proposed progressive structure, which consumes 11,599M memories. Once we turn to ``w/o SOBranch'', the consumption of memory is reduced by 53.67\%, and the training speed increased by 97.16\%. Thus, our progressive structure is useful when training model with GAN. Comparing ``w/o SOBranch'' with PPON (LPIPS values), it naturally demonstrated that SOBranch is beneficial to improve perceptual performance. From Table~\ref{tab:performance}, it can suggest that S-RFN occupies fewer memory footprints and obtains faster training speed than RFN. Besides, the perceptual performance (measured by LPIPS) of S-RFN is significantly improving than RFN evaluated on seven test datasets. Combining Table~\ref{tab:ablation-study-structure} with Table~\ref{tab:performance}, we observe model with GAN (``w/o CRM \& SOBranch'') requires more memories and longer training time. However, the perceptual performance of the model with GAN dramatically boosts than RFN. It means GAN is necessary for our architecture.

Few learnable model parameters (\textbf{1.3M}) complete task migration (\ie from structure-aware to perceptual-aware) well in our work, while ESRGAN~\cite{ESRGAN} uses \textbf{16.7M} to generate perceptual results. We explicitly decompose a task into three subtasks (content, structure, perception). This approach is similar to human painting, first sketching the lines, then adding details. Our topology structure can quickly achieve the migration of similar tasks and infer multiple tasks according to the specific needs.

\begin{figure}[htpb]
	\begin{center}
		\includegraphics[width=0.5\textwidth]{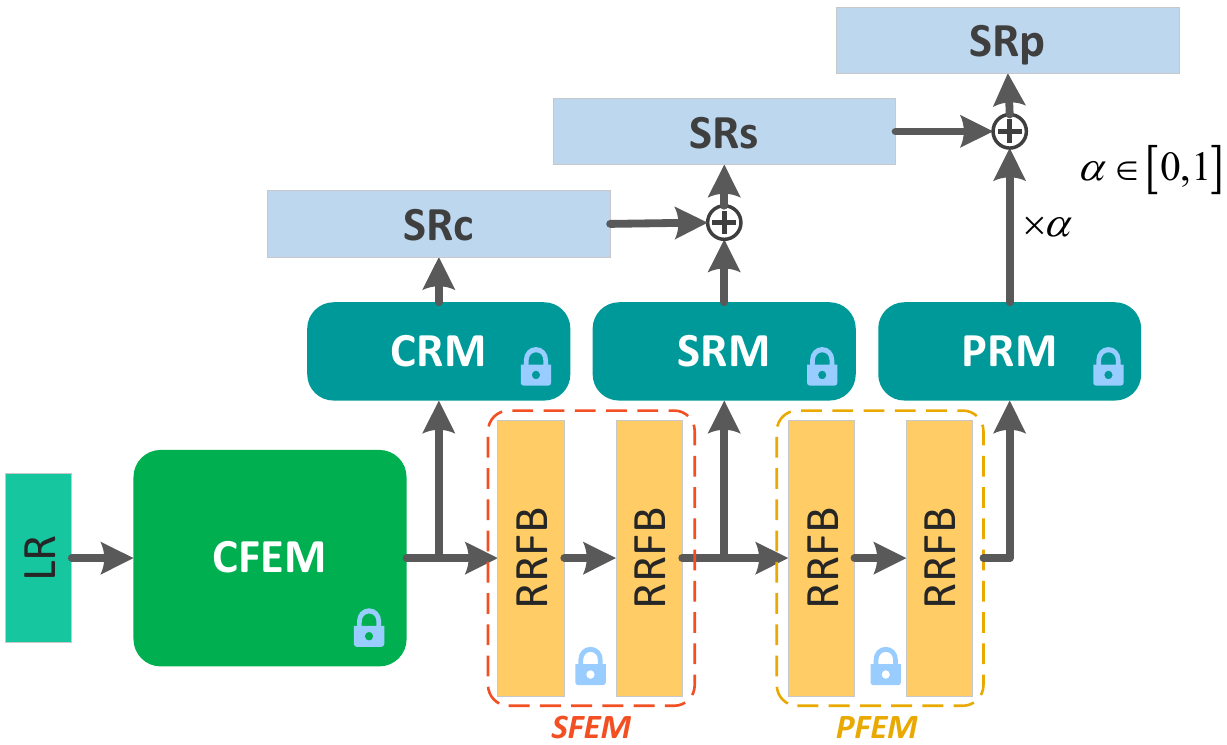}
	\end{center}
	\caption{The inference architecture of our progressive perception-oriented network (PPON).}
	\label{fig:inference-architecture}
\end{figure}

\subsection{Difference to the previous GAN-based methods}
\begin{figure*}[htpb]
	\centering
	\begin{tabular}{ccc}
		$\alpha  = 0.0$ & $\alpha  = 0.6$ & $\alpha  = 1.0$ \\		\includegraphics[width=0.32\textwidth]{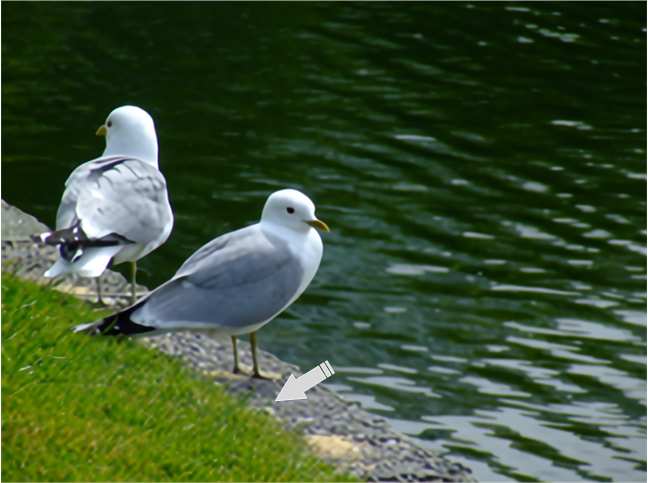} &
		\hspace{-4mm}
		\includegraphics[width=0.32\textwidth]{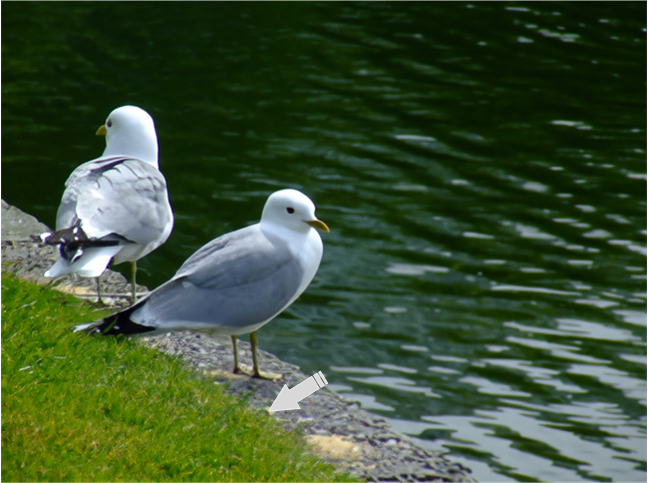} &
		\hspace{-4mm}
		\includegraphics[width=0.32\textwidth]{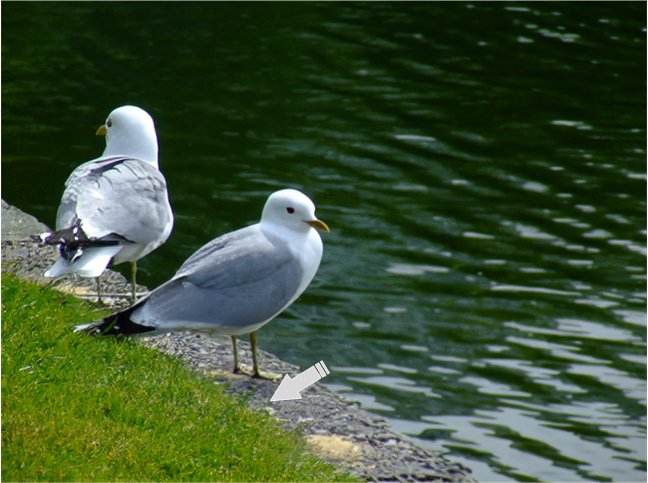} \\
		
		\includegraphics[width=0.32\textwidth]{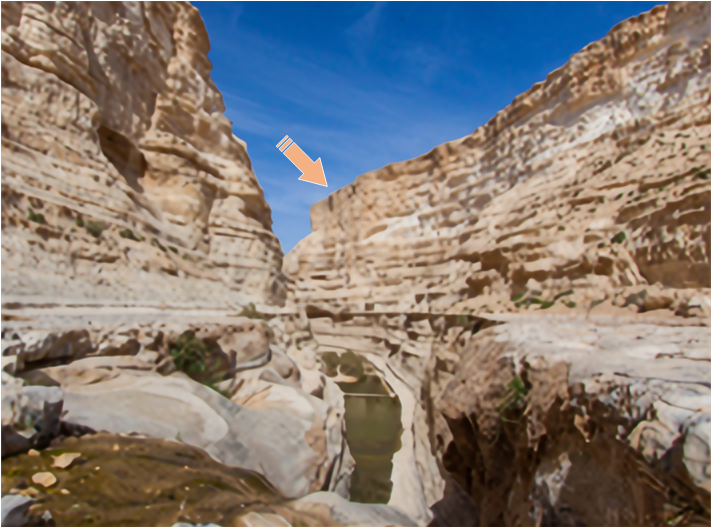} &
		\hspace{-4mm}
		\includegraphics[width=0.32\textwidth]{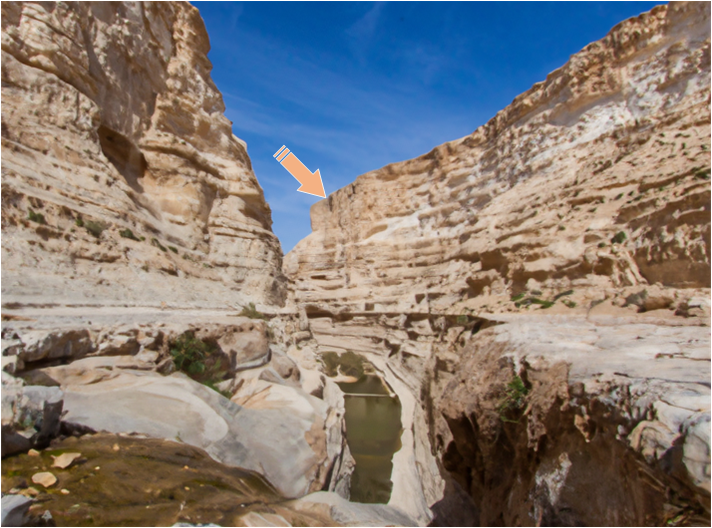} &
		\hspace{-4mm}
		\includegraphics[width=0.32\textwidth]{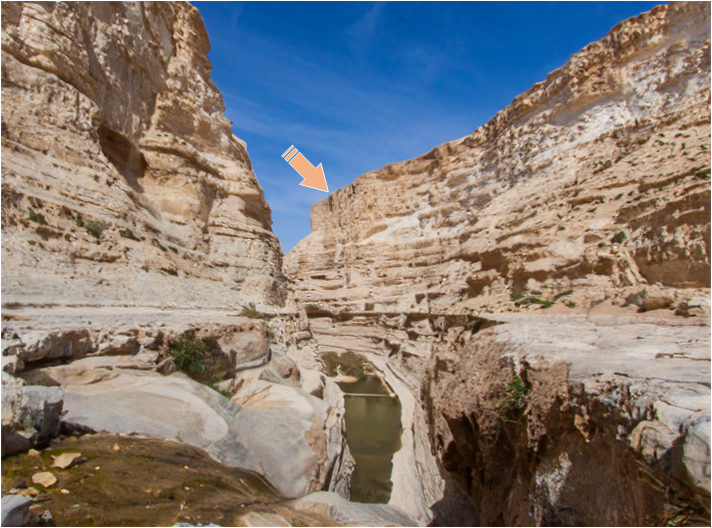} \\
		
	\end{tabular}
	\caption{The perception-distortion trade-off. In the first column, $\alpha  = 0.0$ directly denotes the outputs of SOBranch. Equally, $\alpha  = 1.0$ indicates the results (without any discount) of POBranch. \textbf{Best viewed with zoom-in.}}
	\label{fig:perception-distortion}
\end{figure*}

Unlike the previous perceptual SR methods (~\eg, SRGAN~\cite{SRGAN}, EnhanceNet~\cite{EnhanceNet}, CX~\cite{CX}, and ESRGAN~\cite{ESRGAN}), we employ the progressive strategy to gradually recover the fine-grained high-frequency details without sacrificing the structural information. As shown in Figure~\ref{fig:inference-architecture}, we can obtain images with different perceptions by setting different values to $\alpha $. Now, Equation~\ref{eq:po-branch} can be modified as follows:
\begin{equation}
I_p^\text{SR} = \alpha \cdot{H_\text{PR}}\left( {{F_p}} \right) + I_s^\text{SR} = \alpha \cdot{H_\text{PR}}\left( {{H_\text{PFE}}\left( {{F_s}} \right)} \right) + I_s^\text{SR}.
\end{equation}
We provide an example (see in Figure~\ref{fig:perception-distortion}) to demonstrate the effectiveness of this user-controlled adjustment of SR results.

\subsection{Comparisons with state-of-the-art methods}
\begin{table*}[htpb]
	\caption{Quantitative evaluation results in terms of PSNR and SSIM. \textcolor{red}{\textbf{Red}} and \textcolor{blue}{\underline{blue}} colors indicates the best and second best methods, respectively. Here, S-RFN is the combination of RFN and SOBranch.}
	\label{tab:psnr-ssim-values}
	\begin{center}
		\scalebox{0.7}{
		\begin{tabular}{|l|c|c|c|c|c|c|c|c|c|c|}
			\hline
			\multirow{2}{*}{Method} & \multicolumn{2}{c|}{Set5} & \multicolumn{2}{c|}{Set14} & \multicolumn{2}{c|}{B100} & \multicolumn{2}{c|}{Urban100} & \multicolumn{2}{c|}{Manga109} \\
			\cline{2-11}
			& PSNR & SSIM & PSNR & SSIM & PSNR & SSIM & PSNR & SSIM & PSNR & SSIM \\
			\hline
			\hline
			Bicubic
			& 28.42 & 0.8104
			& 26.00 & 0.7027
			& 25.96 & 0.6675
			& 23.14 & 0.6577
			& 24.89 & 0.7866 \\
			SRCNN~\cite{SRCNN}
			& 30.48 & 0.8628
			& 27.50 & 0.7513
			& 26.90 & 0.7101
			& 24.52 & 0.7221
			& 27.58 & 0.8555 \\
			FSRCNN~\cite{FSRCNN}
			& 30.72 & 0.8660
			& 27.61 & 0.7550
			& 26.98 & 0.7150
			& 24.62 & 0.7280
			& 27.90 & 0.8610 \\
			VDSR~\cite{VDSR}
			& 31.35 & 0.8838
			& 28.01 & 0.7674
			& 27.29 & 0.7251
			& 25.18 & 0.7524
			& 28.87 & 0.8865 \\
			DRCN~\cite{DRCN}
			& 31.53 & 0.8854
			& 28.02 & 0.7670
			& 27.23 & 0.7233
			& 25.14 & 0.7510
			& 28.93 & 0.8854\\
			LapSRN~\cite{LapSRN}
			& 31.54 & 0.8852
			& 28.09 & 0.7700
			& 27.32 & 0.7275
			& 25.21 & 0.7562
			& 29.02 & 0.8900 \\
			MemNet~\cite{MemNet}
			& 31.74 & 0.8893
			& 28.26 & 0.7723
			& 27.40 & 0.7281
			& 25.50 & 0.7630
			& 29.42 & 0.8942 \\
			IDN~\cite{IDN}
			& 31.82 & 0.8903
			& 28.25 & 0.7730
			& 27.41 & 0.7297
			& 25.41 & 0.7632
			& 29.41 & 0.8936 \\
			EDSR~\cite{EDSR}
			& 32.46 & 0.8968
			& 28.80 & 0.7876
			& 27.71 & 0.7420
			& 26.64 & 0.8033
			& 31.02 & 0.9148 \\
			SRMDNF~\cite{SRMDNF}
			& 31.96 & 0.8925
			& 28.35 & 0.7772
			& 27.49 & 0.7337
			& 25.68 & 0.7731
			& 30.09 & 0.9024 \\
			D-DBPN~\cite{DBPN}
			& 32.47 & 0.8980
			& 28.82 & 0.7860
			& 27.72 & 0.7400
			& 26.38 & 0.7946
			& 30.91 & 0.9137 \\
			RDN~\cite{RDN}
			& 32.47 & 0.8990
			& 28.81 & 0.7871
			& 27.72 & 0.7419
			& 26.61 & 0.8028
			& 31.00 & 0.9151 \\
			MSRN~\cite{MSRN}
			& 32.07 & 0.8903
			& 28.60 & 0.7751
			& 27.52 & 0.7273
			& 26.04 & 0.7896
			& 30.17 & 0.9034 \\
			CARN~\cite{CARN}
			& 32.13 & 0.8937
			& 28.60 & 0.7806
			& 27.58 & 0.7349
			& 26.07 & 0.7837
			& 30.47 & 0.9084 \\
			RCAN~\cite{RCAN}
			& 32.63 & 0.9002
			& 28.87 & 0.7889
			& 27.77 & 0.7436
			& 26.82 & 0.8087
			& 31.22 & 0.9173 \\
			SRFBN~\cite{SRFBN}
			& 32.47 & 0.8983
			& 28.81 & 0.7868
			& 27.72 & 0.7409
			& 26.60 & 0.8015
			& 31.15 & 0.9160 \\
			SAN~\cite{SAN}
			& 32.64 & 0.9003
			& \textcolor{blue}{\underline{28.92}} & 0.7888
			& \textcolor{blue}{\underline{27.78}} & 0.7436
			& 26.79 & 0.8068
			& 31.18 & 0.9169 \\
			
			RFN(Ours)
			& \textcolor{red}{\textbf{32.71}} & \textcolor{blue}{\underline{0.9007}}
			& \textcolor{red}{\textbf{28.95}} & \textcolor{blue}{\underline{0.7901}}
			& \textcolor{red}{\textbf{27.83}} & \textcolor{blue}{\underline{0.7449}}
			& \textcolor{red}{\textbf{27.01}} & \textcolor{blue}{\underline{0.8135}}
			& \textcolor{red}{\textbf{31.59}} & \textcolor{blue}{\underline{0.9199}} \\
			S-RFN(Ours)
			& \textcolor{blue}{\underline{32.66}} & \textcolor{red}{\textbf{0.9022}}
			& 28.86 & \textcolor{red}{\textbf{0.7946}}
			& 27.74 & \textcolor{red}{\textbf{0.7515}}
			& \textcolor{blue}{\underline{26.95}} & \textcolor{red}{\textbf{0.8169}} 
			& \textcolor{blue}{\underline{31.51}} & \textcolor{red}{\textbf{0.9211}}
			\\
			\hline
		\end{tabular} }
	\end{center} 
\end{table*}

% PI, PSNR, SSIM, LPIPS
\begin{table*}[htpb]
	\caption{Results on public benchmark datasets, PIRM\_Val, and PIRM\_Test for existing perceptual quality specific methods and our proposed PPON \textcolor{blue}{($\alpha  = 1.0$)}. \textcolor{red}{\textbf{Red}} color indicates the best performance and \textcolor{blue}{\underline{blue}} color indicates the second best performance.}
	\label{tab:pi}
	\begin{center}
		\scalebox{0.58}{
		\begin{tabular}{|c|c|c|c|c|c|c|c|c|c|}
			\hline
			Dataset & Scores & SRGAN~\cite{SRGAN} & ENet~\cite{EnhanceNet} & CX~\cite{CX} & $\text{EPSR}_2$~\cite{EPSR} & $\text{EPSR}_3$~\cite{EPSR} & NatSR~\cite{NatSR}& ESRGAN~\cite{ESRGAN} & PPON (Ours) \\
			\hline
			\hline
			\multirow{4}{*}{Set5} & PSNR & 29.43 & 28.57 & 29.12 & 31.24 & 29.59 & 31.00 & 30.47 & 30.84 \\
			& SSIM & 0.8356 & 0.8103 & 0.8323 & 0.8650 & 0.8415 & 0.8617 & 0.8518 & 0.8561 \\
			& PI & 3.3554 & 2.9261 & 3.2947 & 4.1123 & 3.2571 & 4.1875 & 3.7550 & 3.4590  \\
			& LPIPS & 0.0837 & 0.1014 & 0.0806 & 0.0978 & 0.0889 & 0.0943 & \textcolor{blue}{\underline{0.0748}} & \textcolor{red}{\textbf{0.0664}} \\
			\hline
			\multirow{4}{*}{Set14} & PSNR & 26.12 & 25.77 & 26.06 & 27.77 & 26.36 & 27.53 & 26.28 & 26.97\\
			& SSIM & 0.6958 & 0.6782 & 0.7001 & 0.7440 & 0.7097 & 0.7356 & 0.6984 & 0.7194\\
			& PI & 2.8816 & 3.0176 & 2.7590 & 3.0246 & 2.6981 & 3.1138 & 2.9259 & 2.7741\\
			& LPIPS & 0.1488 & 0.1620 & 0.1452 & 0.1861 & 0.1576 & 0.1765 & \textcolor{blue}{\underline{0.1329}} & \textcolor{red}{\textbf{0.1176}}\\
			\hline
			\multirow{4}{*}{B100} & PSNR & 25.18 & 24.94 & 24.59 & 26.28 & 25.19 & 26.45 & 25.32 & 25.74 \\
			& SSIM & 0.6409 & 0.6266 & 0.6440 & 0.6905 & 0.6468 & 0.6835 & 0.6514 & 0.6684\\
			& PI & 2.3513 & 2.9078 & 2.2501 & 2.7458 & 2.1990 & 2.7746 & 2.4789 & 2.3775\\
			& LPIPS & 0.1843 & 0.2013 & 0.1881 & 0.2474 & 0.2474 & 0.2115 & \textcolor{blue}{\underline{0.1614}} & \textcolor{red}{\textbf{0.1597}}\\
			\hline
			\multirow{4}{*}{PIRM\_Val} & PSNR & N/A & 25.07 & 25.41 & 27.35 & 25.46 & 27.03 & 25.18 & 26.20\\
			& SSIM & N/A & 0.6459 & 0.6747 & 0.7277 & 0.6657 & 0.7199 & 0.6596 & 0.6995\\
			& PI & N/A & 2.6876 & 2.1310 & 2.3880 & 2.0688 & 2.4758 & 2.5550 & 2.2353\\
			& LPIPS & N/A & 0.1667 & 0.1447 & 0.1750 & 0.1869 & 0.1648 & \textcolor{blue}{\underline{0.1443}} & \textcolor{red}{\textbf{0.1194}}\\
			\hline
			\multirow{4}{*}{PIRM\_Test} & PSNR & N/A & 24.95 & 25.31 & 27.04 & 25.35 & 26.95 & 25.04 & 26.01\\
			& SSIM & N/A & 0.6306 & 0.6636 & 0.7068 & 0.6535 & 0.7090 & 0.6454 & 0.6831\\
			& PI & N/A & 2.7232 & 2.1133 & 2.2752 & 2.0131 & 2.3772 & 2.4356 & 2.1511 \\
			& LPIPS & N/A & 0.1776 & \textcolor{blue}{\underline{0.1519}} & 0.1739 & 0.1902 & 0.1712 & 0.1523 & \textcolor{red}{\textbf{0.1273}} \\
			\hline
		\end{tabular} }
	\end{center}
	
\end{table*}

\begin{figure*}[htpb]
	\centering
	\scalebox{0.92}{
	\begin{tabular}{lc}
		\begin{adjustbox}{valign=t}
			\scriptsize
			\begin{tabular}{c}
				\includegraphics[width=0.25\textwidth, height=0.149\textheight]{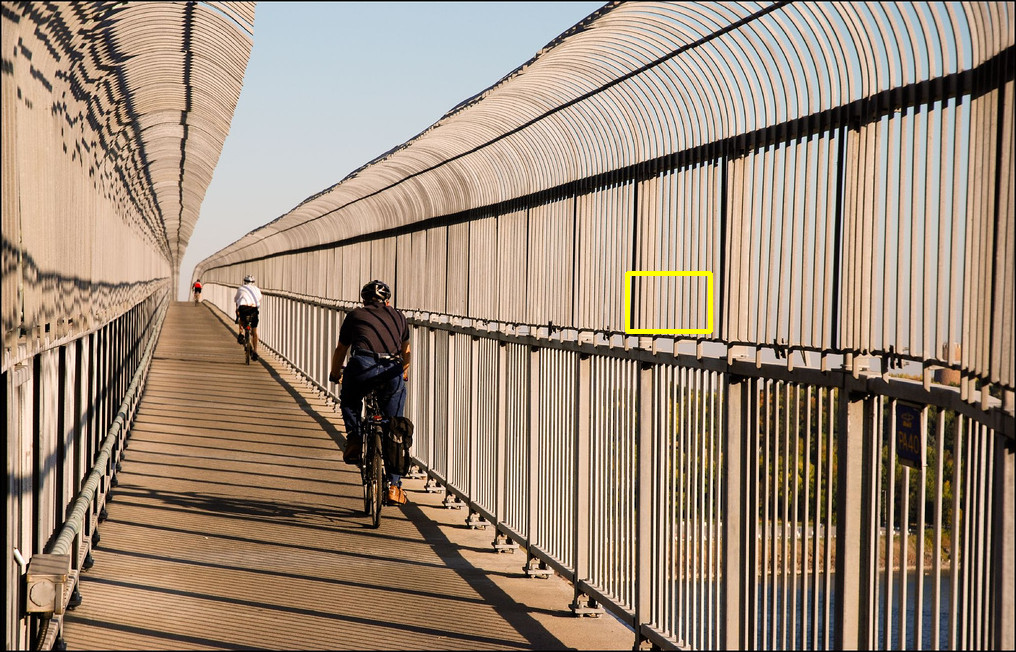} \\
				Urban100 ($4 \times$): \\
				img\_024 \\
			\end{tabular}
		\end{adjustbox}
		\hspace{-3mm}
		\begin{adjustbox}{valign=t}
			\scriptsize
			\begin{tabular}{ccccc}
				\includegraphics[width=0.14\textwidth,height=0.049\textheight]{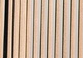} &
				\hspace{-3mm}
				\includegraphics[width=0.14\textwidth,height=0.049\textheight]{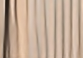} &
				\hspace{-3mm}
				\includegraphics[width=0.14\textwidth,height=0.049\textheight]{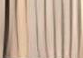} & 
				\hspace{-3mm}
				\includegraphics[width=0.14\textwidth,height=0.049\textheight]{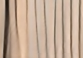} &
				\hspace{-3mm}
				\includegraphics[width=0.14\textwidth,height=0.049\textheight]{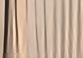} \\
				HR & \hspace{-3mm} VDSR~\cite{VDSR} & \hspace{-3mm} LapSRN~\cite{LapSRN} & \hspace{-3mm} DRRN~\cite{DRRN} & \hspace{-3mm} MemNet~\cite{MemNet} \\
				PSNR/SSIM & \hspace{-3mm} 19.38/0.5925 & \hspace{-3mm} 19.34/0.6037 & \hspace{-3mm} 19.51/0.6161 & \hspace{-3mm} 19.62/0.6179 \\
				\includegraphics[width=0.14\textwidth,height=0.049\textheight]{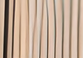} &
				\hspace{-3mm}
				\includegraphics[width=0.14\textwidth,height=0.049\textheight]{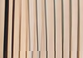} &
				\hspace{-3mm}
				\includegraphics[width=0.14\textwidth,height=0.049\textheight]{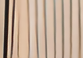} &
				\hspace{-3mm}
				\includegraphics[width=0.14\textwidth,height=0.049\textheight]{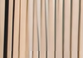} &
				\hspace{-3mm}
				\includegraphics[width=0.14\textwidth,height=0.049\textheight]{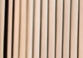} \\
				EDSR~\cite{EDSR} & \hspace{-3mm} RDN~\cite{RDN} & \hspace{-3mm} CARN~\cite{CARN} & \hspace{-3mm} RCAN~\cite{RCAN} & \hspace{-3mm} RFN(Ours) \\
				20.52/0.6826 & \hspace{-3mm} 20.62/0.6827 & \hspace{-3mm} 20.08/0.6449 & \hspace{-3mm} 21.13/0.7119 & \hspace{-3mm} \textbf{21.54/0.7304} \\
			\end{tabular}
			
		\end{adjustbox}
		\\
		\begin{adjustbox}{valign=t}
			\scriptsize
			\begin{tabular}{c}
				\includegraphics[width=0.25\textwidth, height=0.149\textheight]{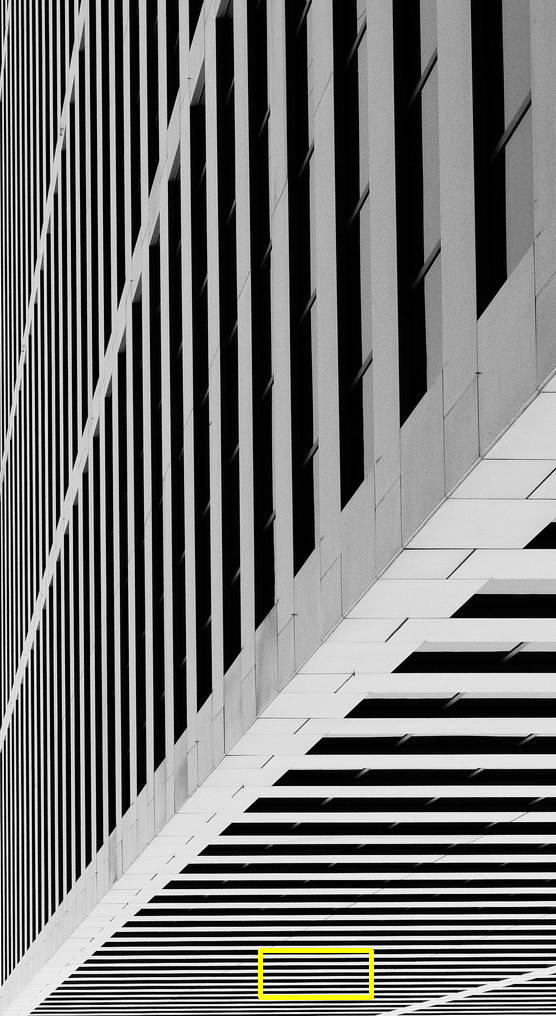} \\
				Urban100 ($4 \times$): \\
				img\_011 \\
			\end{tabular}
		\end{adjustbox}
		\hspace{-3mm}
		\begin{adjustbox}{valign=t}
			\scriptsize
			\begin{tabular}{ccccc}
				\includegraphics[width=0.14\textwidth,height=0.049\textheight]{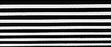} &
				\hspace{-3mm}
				\includegraphics[width=0.14\textwidth,height=0.049\textheight]{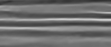} &
				\hspace{-3mm}
				\includegraphics[width=0.14\textwidth,height=0.049\textheight]{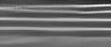} & 
				\hspace{-3mm}
				\includegraphics[width=0.14\textwidth,height=0.049\textheight]{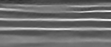} &
				\hspace{-3mm}
				\includegraphics[width=0.14\textwidth,height=0.049\textheight]{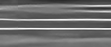} \\
				HR & \hspace{-3mm} VDSR~\cite{VDSR} & \hspace{-3mm} LapSRN~\cite{LapSRN} & \hspace{-3mm} DRRN~\cite{DRRN} & \hspace{-3mm} MemNet~\cite{MemNet} \\
				PSNR/SSIM & \hspace{-3mm} 17.29/0.7565 & \hspace{-3mm} 17.21/0.7568 & \hspace{-3mm} 17.12/0.7660 & \hspace{-3mm} 17.29/0.7773 \\
				\includegraphics[width=0.14\textwidth,height=0.049\textheight]{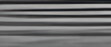} &
				\hspace{-3mm}
				\includegraphics[width=0.14\textwidth,height=0.049\textheight]{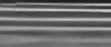} &
				\hspace{-3mm}
				\includegraphics[width=0.14\textwidth,height=0.049\textheight]{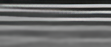} &
				\hspace{-3mm}
				\includegraphics[width=0.14\textwidth,height=0.049\textheight]{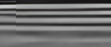} &
				\hspace{-3mm}
				\includegraphics[width=0.14\textwidth,height=0.049\textheight]{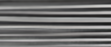} \\
				EDSR~\cite{EDSR} &\hspace{-3mm} RDN~\cite{RDN} &\hspace{-3mm} CARN~\cite{CARN} &\hspace{-3mm} RCAN~\cite{RCAN} &\hspace{-3mm} RFN(Ours) \\
				18.42/0.8225 &\hspace{-3mm} 17.85/0.8003 &\hspace{-3mm} 17.71/0.7986 &\hspace{-3mm} 18.42/0.8224 &\hspace{-3mm} \textbf{19.17/0.8485} \\
			\end{tabular}
		\end{adjustbox}
		\\
		\begin{adjustbox}{valign=t}
			\scriptsize
			\begin{tabular}{c}
				\includegraphics[width=0.25\textwidth, height=0.149\textheight]{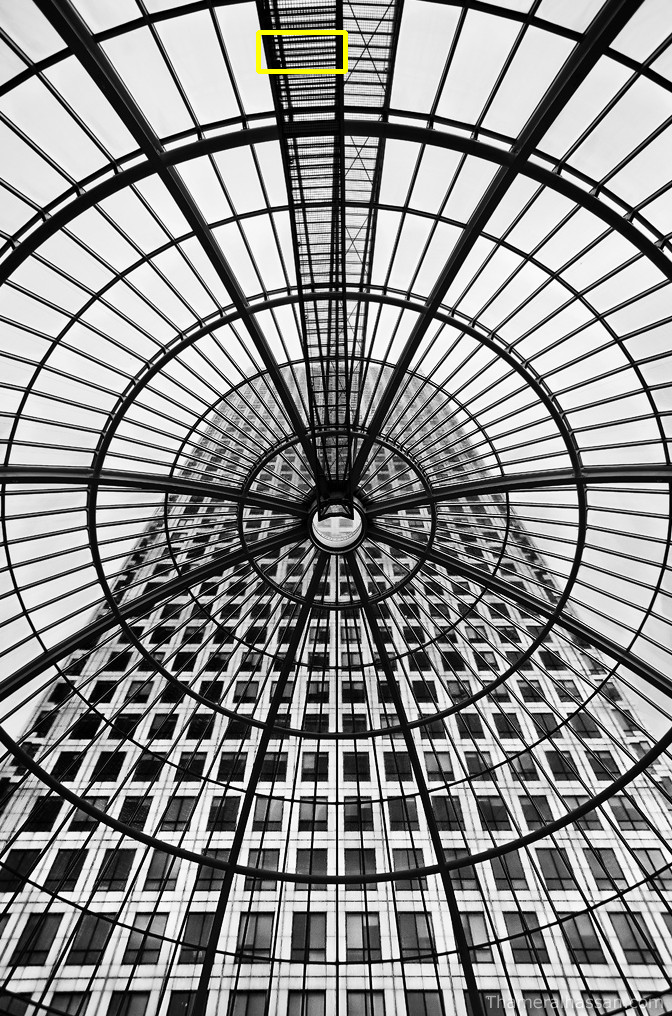} \\
				Urban100 ($4 \times$): \\
				img\_072 \\
			\end{tabular}
		\end{adjustbox}
		\hspace{-4mm}
		\begin{adjustbox}{valign=t}
			\scriptsize
			\begin{tabular}{ccccc}
				\includegraphics[width=0.14\textwidth,height=0.049\textheight]{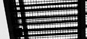} &
				\hspace{-3mm}
				\includegraphics[width=0.14\textwidth,height=0.049\textheight]{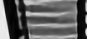} &
				\hspace{-3mm}
				\includegraphics[width=0.14\textwidth,height=0.049\textheight]{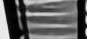} & 
				\hspace{-3mm}
				\includegraphics[width=0.14\textwidth,height=0.049\textheight]{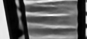} &
				\hspace{-3mm}
				\includegraphics[width=0.14\textwidth,height=0.049\textheight]{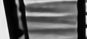} \\
				HR & VDSR~\cite{VDSR} & LapSRN~\cite{LapSRN} & DRRN~\cite{DRRN} & MemNet~\cite{MemNet} \\
				PSNR/SSIM & 19.19/0.7713 & 19.33/0.7836 & 19.91/0.8056 & 20.01/0.8099 \\
				\includegraphics[width=0.14\textwidth,height=0.049\textheight]{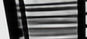} &
				\hspace{-3mm}
				\includegraphics[width=0.14\textwidth,height=0.049\textheight]{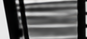} &
				\hspace{-3mm}
				\includegraphics[width=0.14\textwidth,height=0.049\textheight]{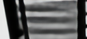} &
				\hspace{-3mm}
				\includegraphics[width=0.14\textwidth,height=0.049\textheight]{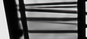} &
				\hspace{-3mm}
				\includegraphics[width=0.14\textwidth,height=0.049\textheight]{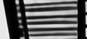} \\
				EDSR~\cite{EDSR} &\hspace{-3mm} RDN~\cite{RDN} &\hspace{-3mm} CARN~\cite{CARN} &\hspace{-3mm} RCAN~\cite{RCAN} &\hspace{-3mm} RFN(Ours) \\
				21.24/0.8573 &\hspace{-3mm} 21.12/0.8540 &\hspace{-3mm} 20.44/0.8297 &\hspace{-3mm} 21.46/0.8656 &\hspace{-3mm} \textbf{21.91/0.8745} \\
			\end{tabular}	
		\end{adjustbox}
		\\
		\begin{adjustbox}{valign=t}
			\scriptsize
			\begin{tabular}{c}
				\includegraphics[width=0.25\textwidth, height=0.149\textheight]{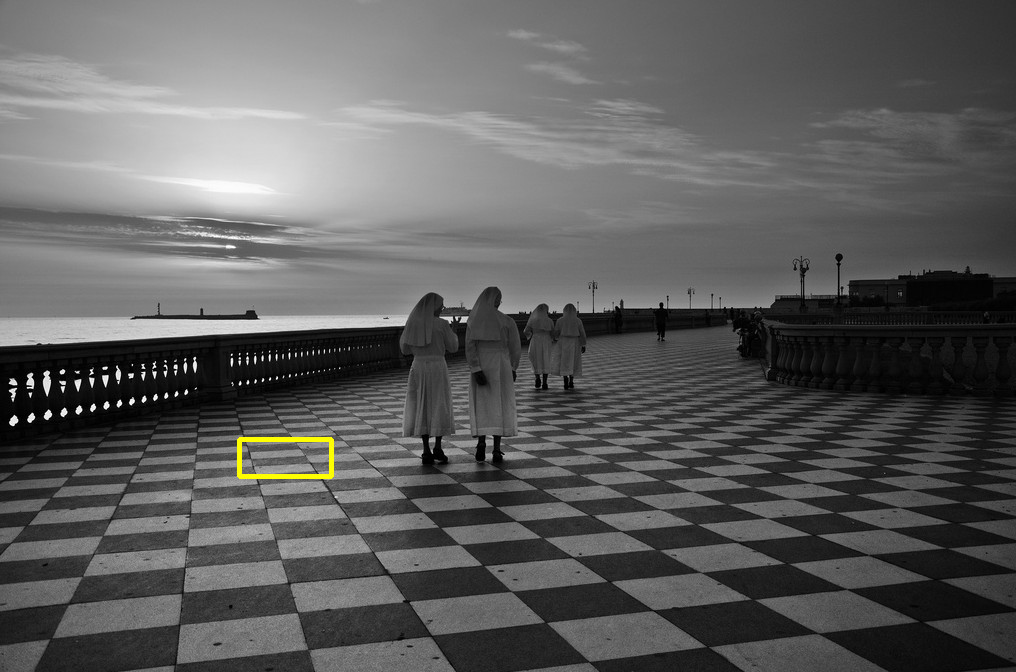} \\
				Urban100 ($4 \times$): \\
				img\_028 \\
			\end{tabular}
		\end{adjustbox}
		\hspace{-4mm}
		\begin{adjustbox}{valign=t}
			\scriptsize
			\begin{tabular}{ccccc}
				\includegraphics[width=0.14\textwidth,height=0.049\textheight]{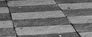} &
				\hspace{-3mm}
				\includegraphics[width=0.14\textwidth,height=0.049\textheight]{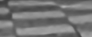} &
				\hspace{-3mm}
				\includegraphics[width=0.14\textwidth,height=0.049\textheight]{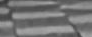} & 
				\hspace{-3mm}
				\includegraphics[width=0.14\textwidth,height=0.049\textheight]{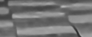} &
				\hspace{-3mm}
				\includegraphics[width=0.14\textwidth,height=0.049\textheight]{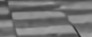} \\
				HR & VDSR~\cite{VDSR} & LapSRN~\cite{LapSRN} & DRRN~\cite{DRRN} & MemNet~\cite{MemNet} \\
				PSNR/SSIM & 31.01/0.8683 & 30.97/0.8707 & 31.10/0.8702 & 31.10/0.8690 \\
				\includegraphics[width=0.14\textwidth,height=0.049\textheight]{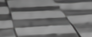} &
				\hspace{-3mm}
				\includegraphics[width=0.14\textwidth,height=0.049\textheight]{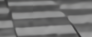} &
				\hspace{-3mm}
				\includegraphics[width=0.14\textwidth,height=0.049\textheight]{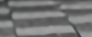} &
				\hspace{-3mm}
				\includegraphics[width=0.14\textwidth,height=0.049\textheight]{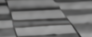} &
				\hspace{-3mm}
				\includegraphics[width=0.14\textwidth,height=0.049\textheight]{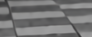} \\
				EDSR~\cite{EDSR} &\hspace{-3mm} RDN~\cite{RDN} &\hspace{-3mm} CARN~\cite{CARN} &\hspace{-3mm} RCAN~\cite{RCAN} &\hspace{-3mm} RFN(Ours) \\
				31.95/0.8861 &\hspace{-3mm} 31.85/0.8833 &\hspace{-3mm} 31.42/0.8769 &\hspace{-3mm} 31.96/0.8857 &\hspace{-3mm} \textbf{32.37/0.8915} \\
			\end{tabular}	
		\end{adjustbox}
		\\
		\begin{adjustbox}{valign=t}
			\scriptsize
			\begin{tabular}{c}
				\includegraphics[width=0.25\textwidth, height=0.149\textheight]{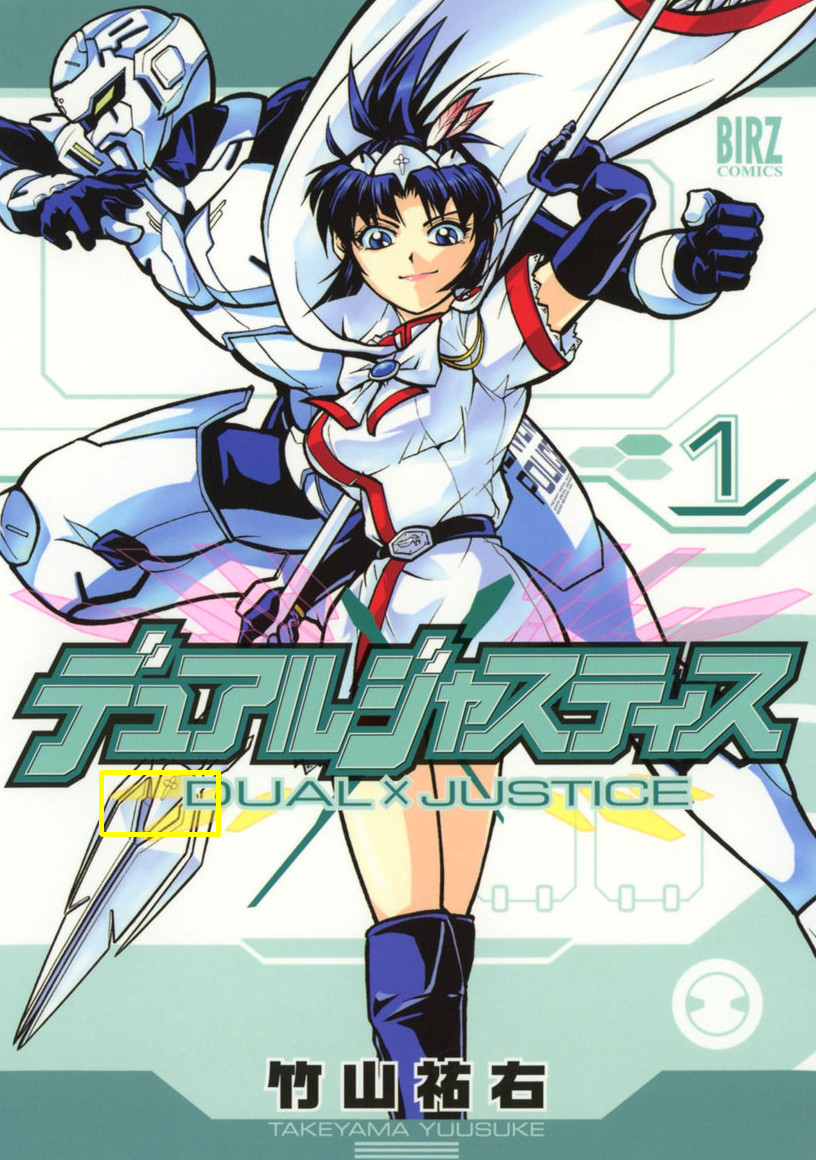} \\
				Manga109 ($4 \times$): \\
				DualJustice \\
			\end{tabular}
		\end{adjustbox}
		\hspace{-3mm}
		\begin{adjustbox}{valign=t}
			\scriptsize
			\begin{tabular}{ccccc}
				\includegraphics[width=0.14\textwidth,height=0.049\textheight]{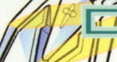} &
				\hspace{-3mm}
				\includegraphics[width=0.14\textwidth,height=0.049\textheight]{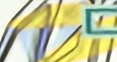} &
				\hspace{-3mm}
				\includegraphics[width=0.14\textwidth,height=0.049\textheight]{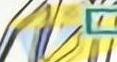} & 
				\hspace{-3mm}
				\includegraphics[width=0.14\textwidth,height=0.049\textheight]{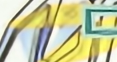} &
				\hspace{-3mm}
				\includegraphics[width=0.14\textwidth]{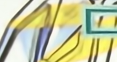} \\
				HR & \hspace{-3mm} VDSR~\cite{VDSR} & \hspace{-3mm} LapSRN~\cite{LapSRN} & \hspace{-3mm} DRRN~\cite{DRRN} & \hspace{-3mm} MemNet~\cite{MemNet} \\
				PSNR/SSIM & \hspace{-3mm} 28.04/0.9201 & \hspace{-3mm} 28.30/0.9239 & \hspace{-3mm} 29.15/0.9356 & \hspace{-3mm} 29.28/0.9378 \\
				\includegraphics[width=0.14\textwidth,height=0.049\textheight]{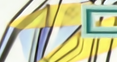} &
				\hspace{-3mm}
				\includegraphics[width=0.14\textwidth,height=0.049\textheight]{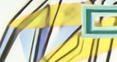} &
				\hspace{-3mm}
				\includegraphics[width=0.14\textwidth,height=0.049\textheight]{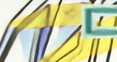} &
				\hspace{-3mm}
				\includegraphics[width=0.14\textwidth,height=0.049\textheight]{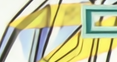} &
				\hspace{-3mm}
				\includegraphics[width=0.14\textwidth,height=0.049\textheight]{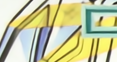} \\
				EDSR~\cite{EDSR} & \hspace{-3mm} RDN~\cite{RDN} & \hspace{-3mm} CARN~\cite{CARN} & \hspace{-3mm} RCAN~\cite{RCAN} & \hspace{-3mm} RFN(Ours) \\
				31.55/0.9626 & \hspace{-3mm} 31.41/0.9617 & \hspace{-3mm} 30.46/0.9516 & \hspace{-3mm} 32.39/0.9688 & \hspace{-3mm} \textbf{33.01/0.9724} \\
			\end{tabular}
		\end{adjustbox}
	\end{tabular} }
	\caption{Visual comparisons for $4 \times$ SR with RFN on Urban100 and Manga109 datasets.}
	\label{fig:compare-PSNR}
\end{figure*}

We compare our RFN with $16$ state-of-the-art methods: SRCNN~\cite{SRCNN,SRCNN-Ex}, FSRCNN~\cite{FSRCNN}, VDSR~\cite{VDSR}, DRCN~\cite{DRCN}, LapSRN~\cite{LapSRN}, MemNet~\cite{MemNet}, IDN~\cite{IDN}, EDSR~\cite{EDSR}, SRMDNF~\cite{SRMDNF}, D-DBPN~\cite{DBPN}, RDN~\cite{RDN}, MSRN~\cite{MSRN}, CARN~\cite{CARN}, RCAN~\cite{RCAN}, SAN~\cite{SAN}, and SRFBN~\cite{SRFBN}. Table~\ref{tab:psnr-ssim-values} shows quantitative comparisons for $\times 4$ SR. It can be seen that our RFN performs the best in terms of PSNR on all the datasets. The proposed S-RFN shows significant advantages of SSIM. In Figure~\ref{fig:compare-PSNR}, we present visual comparisons on different datasets. For image ``img\_011'', we observe that most of the compared methods cannot recover the lines and suffer from blurred artifacts. In contrast, our RFN can slightly alleviate this phenomenon and restore more details. 

\begin{figure*}[htpb]
	\centering
	\scalebox{0.58}{
		\begin{tabular}{lc}
			% baboon
			\begin{adjustbox}{valign=t}
				\scriptsize
				\begin{tabular}{c}
					\includegraphics[width=0.2\textwidth, height=0.127\textheight]{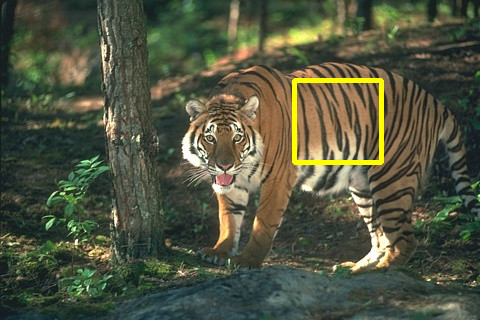} \\
					BSD100 ($4 \times$): \\
					108005 \\
				\end{tabular}
			\end{adjustbox}
			\hspace{-3mm}
			\begin{adjustbox}{valign=t}
				\scriptsize
				\begin{tabular}{cccccccc}
					\includegraphics[width=0.174\textwidth, height=0.126\textheight]{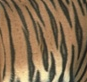} &
					\hspace{-3mm}
					\includegraphics[width=0.174\textwidth, height=0.126\textheight]{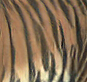} &
					\hspace{-3mm}
					\includegraphics[width=0.174\textwidth, height=0.126\textheight]{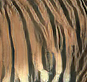} &
					\hspace{-3mm}
					\includegraphics[width=0.174\textwidth, height=0.126\textheight]{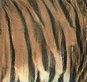} &
					\hspace{-3mm}
					\includegraphics[width=0.174\textwidth, height=0.126\textheight]{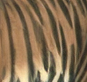} &
					\hspace{-3mm}
					\includegraphics[width=0.174\textwidth, height=0.126\textheight]{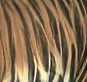} &
					\hspace{-3mm}
					\includegraphics[width=0.174\textwidth, height=0.126\textheight]{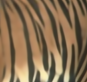} &
					\hspace{-3mm}
					\includegraphics[width=0.174\textwidth, height=0.126\textheight]{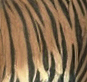} \\
					HR & \hspace{-3mm} SRGAN~\cite{SRGAN} & \hspace{-3mm} ENet~\cite{EnhanceNet} & \hspace{-3mm} CX~\cite{CX} & \hspace{-3mm} $\text{EPSR}_2$~\cite{EPSR} & \hspace{-3mm} ESRGAN~\cite{ESRGAN} & \hspace{-3mm} S-RFN(Ours) & \hspace{-3mm} PPON(Ours)\\
					
					PSNR/LPIPS & \hspace{-3mm} 23.70/0.2471 & \hspace{-3mm} 23.27/0.2547 & \hspace{-3mm} 23.82/0.1999 & \hspace{-3mm} 25.72/0.2224 & \hspace{-3mm} 23.54/0.1806 & \hspace{-3mm} 26.48/0.2998 & \hspace{-3mm} 24.39/\textbf{0.1546} \\
				\end{tabular}
			\end{adjustbox}
			\\
			
			\begin{adjustbox}{valign=t}
				\scriptsize
				\begin{tabular}{c}
					\includegraphics[width=0.2\textwidth, height=0.127\textheight]{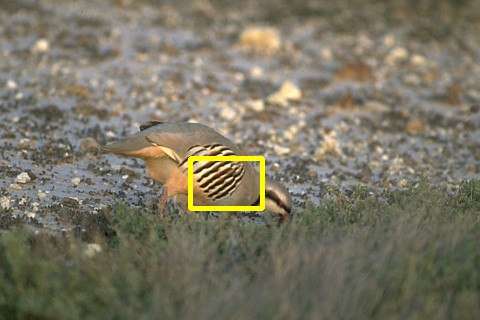} \\
					BSD100 ($4 \times$): \\
					8023 \\
				\end{tabular}
			\end{adjustbox}
			\hspace{-3mm}
			\begin{adjustbox}{valign=t}
				\scriptsize
				\begin{tabular}{cccccccc}
					\includegraphics[width=0.174\textwidth,height=0.126\textheight]{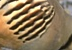} &
					\hspace{-3mm}
					\includegraphics[width=0.174\textwidth,height=0.126\textheight]{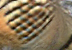} &
					\hspace{-3mm}
					\includegraphics[width=0.174\textwidth,height=0.126\textheight]{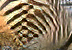} &
					\hspace{-3mm}
					\includegraphics[width=0.174\textwidth,height=0.126\textheight]{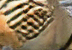} &
					\hspace{-3mm}
					\includegraphics[width=0.174\textwidth,height=0.126\textheight]{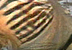} &
					\hspace{-3mm}
					\includegraphics[width=0.174\textwidth,height=0.126\textheight]{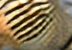} &
					\hspace{-3mm}
					\includegraphics[width=0.174\textwidth,height=0.126\textheight]{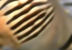} &
					\hspace{-3mm}
					\includegraphics[width=0.174\textwidth,height=0.126\textheight]{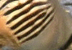} \\
					HR & \hspace{-3mm} SRGAN~\cite{SRGAN} & \hspace{-3mm} ENet~\cite{EnhanceNet} & \hspace{-3mm} CX~\cite{CX} & \hspace{-3mm} $\text{EPSR}_2$~\cite{EPSR} & \hspace{-3mm} ESRGAN~\cite{ESRGAN} & \hspace{-3mm} S-RFN(Ours) & \hspace{-3mm} PPON(Ours)\\
					
					PSNR/LPIPS & \hspace{-3mm} 28.78/0.1355 & \hspace{-3mm} 25.71/0.3490 & \hspace{-3mm} 27.69/0.1610 & \hspace{-3mm} 29.97/0.2064 & \hspace{-3mm} 29.58/0.1554 & \hspace{-3mm} 31.17/0.2807 & \hspace{-3mm} 29.78/\textbf{0.1332} \\
				\end{tabular}
			\end{adjustbox}
			\\
			
			% PIRM_Val 86
			\begin{adjustbox}{valign=t}
				\scriptsize
				\begin{tabular}{c}
					\includegraphics[width=0.2\textwidth, height=0.122\textheight]{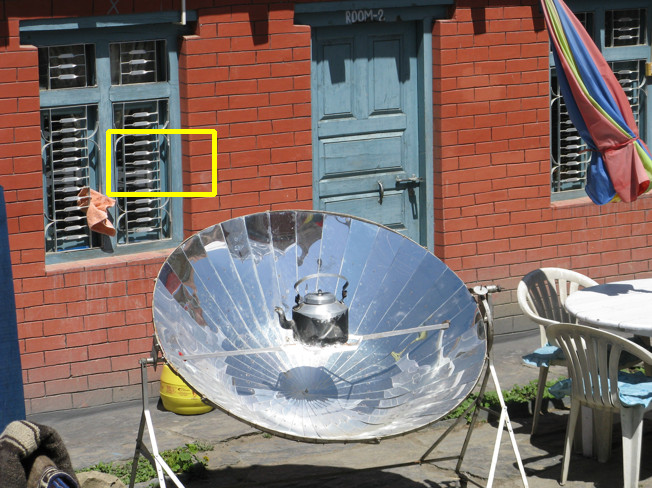} \\
					PIRM\_Val ($4 \times$): \\
					86 \\
				\end{tabular}
			\end{adjustbox}
			\hspace{-3mm}
			\begin{adjustbox}{valign=t}
				\scriptsize
				\begin{tabular}{cccccccc}
					\includegraphics[width=0.174\textwidth, height=0.122\textheight]{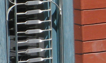} &
					\hspace{-3mm}
					\includegraphics[width=0.174\textwidth, height=0.122\textheight]{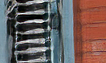} &
					\hspace{-3mm}
					\includegraphics[width=0.174\textwidth, height=0.122\textheight]{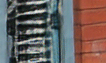} &
					\hspace{-3mm}
					\includegraphics[width=0.174\textwidth, height=0.122\textheight]{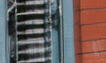} &
					\hspace{-3mm}
					
					\includegraphics[width=0.174\textwidth, height=0.122\textheight]{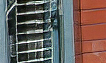} &
					\hspace{-3mm}
					\includegraphics[width=0.174\textwidth, height=0.122\textheight]{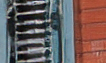} &
					\hspace{-3mm}
					\includegraphics[width=0.174\textwidth, height=0.122\textheight]{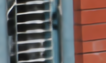} &
					\hspace{-3mm}
					\includegraphics[width=0.174\textwidth, height=0.122\textheight]{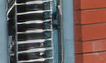} \\
					HR & \hspace{-3mm} ENet~\cite{EnhanceNet} & \hspace{-3mm} CX~\cite{CX} & \hspace{-3mm} $\text{EPSR}_2$~\cite{EPSR} &
					\hspace{-3mm} ESRGAN~\cite{ESRGAN} & \hspace{-3mm} SuperSR~\cite{ESRGAN} & \hspace{-3mm} S-RFN(Ours) & \hspace{-3mm} PPON(Ours) \\
					PSNR/LPIPS & \hspace{-3mm} 24.16/0.1276 & \hspace{-3mm} 24.88/0.1248 & \hspace{-3mm} 26.79/0.1031 & \hspace{-3mm} 24.02/0.1189 & \hspace{-3mm} 25.03/0.1353 & \hspace{-3mm} 28.47/0.1352 & \hspace{-3mm} 26.32/\textbf{0.0773} \\
				\end{tabular}
			\end{adjustbox}
			\\
			\begin{adjustbox}{valign=t}
				\scriptsize
				\begin{tabular}{c}
					\includegraphics[width=0.2\textwidth, height=0.116\textheight]{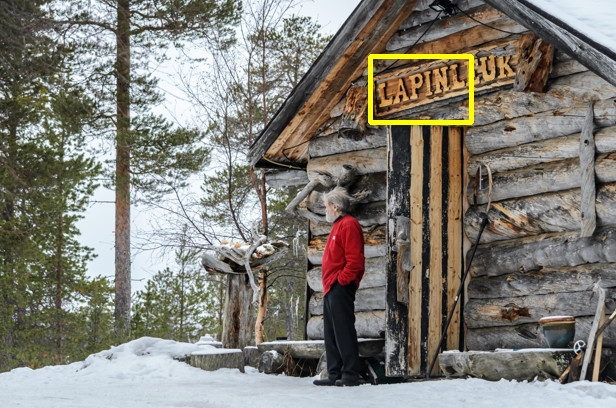} \\
					PIRM\_Test ($4 \times$): \\
					223 \\
				\end{tabular}
			\end{adjustbox}
			\hspace{-3mm}
			\begin{adjustbox}{valign=t}
				\scriptsize
				\begin{tabular}{cccccccc}
					\includegraphics[width=0.174\textwidth,height=0.115\textheight]{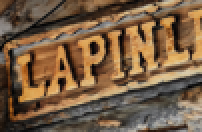} &
					\hspace{-3mm}
					\includegraphics[width=0.174\textwidth,height=0.115\textheight]{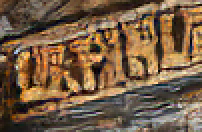} &
					\hspace{-3mm}
					\includegraphics[width=0.174\textwidth,height=0.115\textheight]{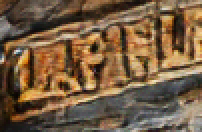} &
					\hspace{-3mm}
					\includegraphics[width=0.174\textwidth,height=0.115\textheight]{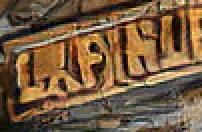} &
					\hspace{-3mm}
					\includegraphics[width=0.174\textwidth,height=0.115\textheight]{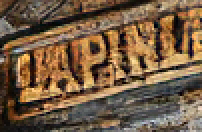} &
					\hspace{-3mm}
					\includegraphics[width=0.174\textwidth,height=0.115\textheight]{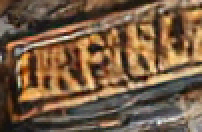} &
					\hspace{-3mm}
					\includegraphics[width=0.174\textwidth,height=0.115\textheight]{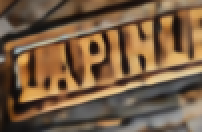} &
					\hspace{-3mm}
					\includegraphics[width=0.174\textwidth,height=0.115\textheight]{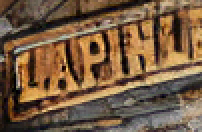} \\
					HR & \hspace{-3mm} ENet~\cite{EnhanceNet} & \hspace{-3mm} CX~\cite{CX} & \hspace{-3mm} $\text{EPSR}_3$~\cite{EPSR} & \hspace{-3mm} ESRGAN~\cite{ESRGAN} & \hspace{-3mm} SuperSR~\cite{ESRGAN} & \hspace{-3mm} S-RFN(Ours) & \hspace{-3mm} PPON(Ours) \\
					PSNR/LPIPS & \hspace{-3mm} 19.80/0.1756 & \hspace{-3mm} 20.64/0.1552 & \hspace{-3mm} 20.15/0.1797 & \hspace{-3mm} 18.96/0.2128 & \hspace{-3mm} 20.43/0.1710 & \hspace{-3mm} 23.11/0.2915 & \hspace{-3mm} 20.22/\textbf{0.1466} \\
				\end{tabular}
			\end{adjustbox}
			\\
			\begin{adjustbox}{valign=t}
				\scriptsize
				\begin{tabular}{c}
					\includegraphics[width=0.2\textwidth, height=0.116\textheight]{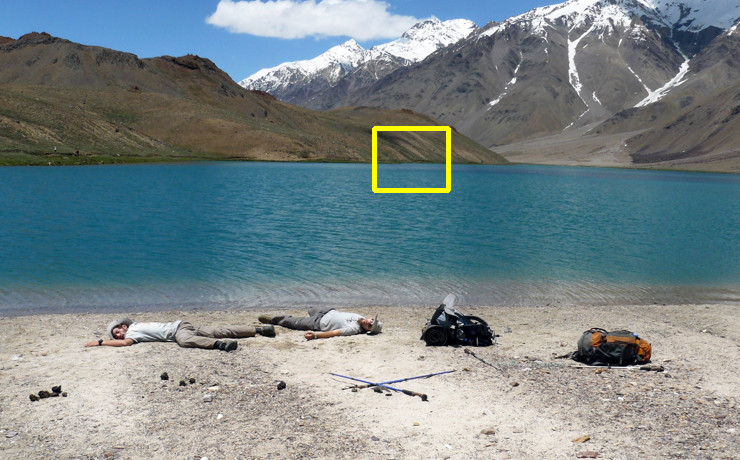} \\
					PIRM\_Val ($4 \times$): \\
					43 \\
				\end{tabular}
			\end{adjustbox}
			\hspace{-3mm}
			\begin{adjustbox}{valign=t}
				\scriptsize
				\begin{tabular}{cccccccc}
					\includegraphics[width=0.174\textwidth,height=0.115\textheight]{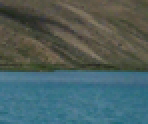} &
					\hspace{-3mm}
					\includegraphics[width=0.174\textwidth,height=0.115\textheight]{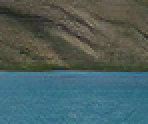} &
					\hspace{-3mm}
					\includegraphics[width=0.174\textwidth,height=0.115\textheight]{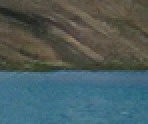} &
					\hspace{-3mm}
					\includegraphics[width=0.174\textwidth,height=0.115\textheight]{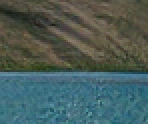} &
					\hspace{-3mm}					
					\includegraphics[width=0.174\textwidth,height=0.115\textheight]{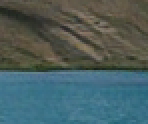} &
					\hspace{-3mm}
					\includegraphics[width=0.174\textwidth,height=0.115\textheight]{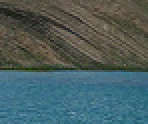} &
					\hspace{-3mm}
					\includegraphics[width=0.174\textwidth,height=0.115\textheight]{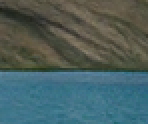} &
					\hspace{-3mm}
					\includegraphics[width=0.174\textwidth,height=0.115\textheight]{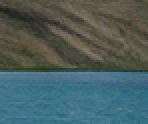} \\
					
					HR & \hspace{-3mm} ENet~\cite{EnhanceNet} & \hspace{-3mm} CX~\cite{CX} & \hspace{-3mm} EPSR3~\cite{EPSR} & 
					SuperSR~\cite{ESRGAN} & \hspace{-3mm} ESRGAN~\cite{ESRGAN} & \hspace{-3mm} PPON\_128 (Ours) & \hspace{-3mm} PPON (Ours) \\
					
					PSNR/LPIPS & \hspace{-3mm} 24.77/0.1560 & \hspace{-3mm} 25.55/0.1466 & \hspace{-3mm} 24.53/0.2305 &
					25.12/0.1421 & \hspace{-3mm} 23.80/0.1861 & \hspace{-3mm} 24.57/0.1320 & \hspace{-3mm} 24.87/\textbf{0.1256} \\
				\end{tabular}
			\end{adjustbox}
			\\
			\begin{adjustbox}{valign=t}
				\scriptsize
				\begin{tabular}{c}
					\includegraphics[width=0.2\textwidth, height=0.116\textheight]{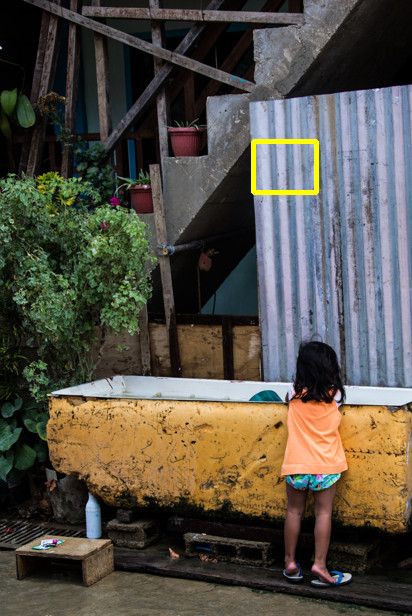} \\
					PIRM\_Val ($4 \times$): \\
					64 \\
				\end{tabular}
			\end{adjustbox}
			\hspace{-3mm}
			\begin{adjustbox}{valign=t}
				\scriptsize
				\begin{tabular}{cccccccc}
					\includegraphics[width=0.174\textwidth,height=0.115\textheight]{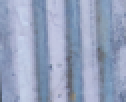} &
					\hspace{-3mm}
					\includegraphics[width=0.174\textwidth,height=0.115\textheight]{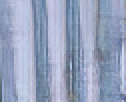} &
					\hspace{-3mm}
					\includegraphics[width=0.174\textwidth,height=0.115\textheight]{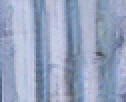} &
					\hspace{-3mm}
					\includegraphics[width=0.174\textwidth,height=0.115\textheight]{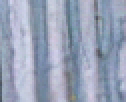} &
					\hspace{-3mm}					
					\includegraphics[width=0.174\textwidth,height=0.115\textheight]{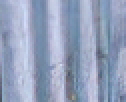} &
					\hspace{-3mm}
					\includegraphics[width=0.174\textwidth,height=0.115\textheight]{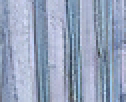} &
					\hspace{-3mm}
					\includegraphics[width=0.174\textwidth,height=0.115\textheight]{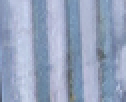} &
					\hspace{-3mm}
					\includegraphics[width=0.174\textwidth,height=0.115\textheight]{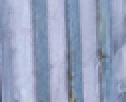} \\
					
					HR & \hspace{-3mm} ENet~\cite{EnhanceNet} & \hspace{-3mm} CX~\cite{CX} & \hspace{-3mm} EPSR3~\cite{EPSR} & 
					SuperSR~\cite{ESRGAN} & \hspace{-3mm} ESRGAN~\cite{ESRGAN} & \hspace{-3mm} PPON\_128 (Ours) & \hspace{-3mm} PPON (Ours) \\
					
					PSNR/LPIPS & \hspace{-3mm} 25.04/0.1429 & \hspace{-3mm} 25.60/0.1301 & \hspace{-3mm} 25.36/0.1597 &
					25.60/0.1468 & \hspace{-3mm} 24.56/0.1513 & \hspace{-3mm} 25.84/0.1092 & \hspace{-3mm} 25.91/\textbf{0.1067} \\
				\end{tabular}
			\end{adjustbox}
			\\
			\begin{adjustbox}{valign=t}
				\scriptsize
				\begin{tabular}{c}
					\includegraphics[width=0.2\textwidth, height=0.116\textheight]{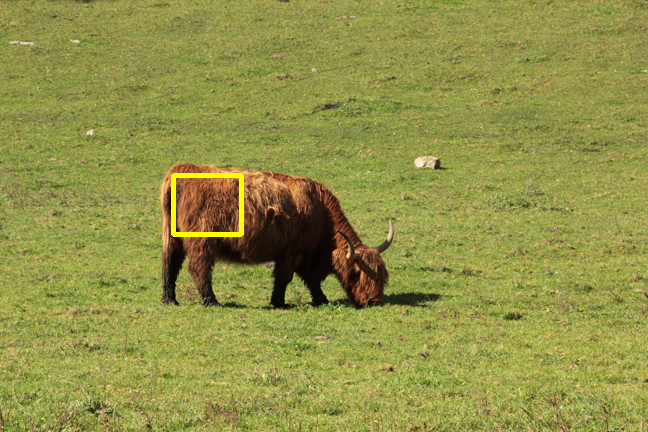} \\
					PIRM\_Val ($4 \times$): \\
					84 \\
				\end{tabular}
			\end{adjustbox}
			\hspace{-3mm}
			\begin{adjustbox}{valign=t}
				\scriptsize
				\begin{tabular}{cccccccc}
					\includegraphics[width=0.174\textwidth,height=0.115\textheight]{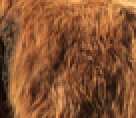} &
					\hspace{-3mm}
					\includegraphics[width=0.174\textwidth,height=0.115\textheight]{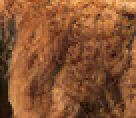} &
					\hspace{-3mm}
					\includegraphics[width=0.174\textwidth,height=0.115\textheight]{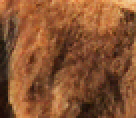} &
					\hspace{-3mm}
					\includegraphics[width=0.174\textwidth,height=0.115\textheight]{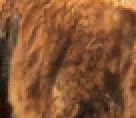} &
					\hspace{-3mm}					
					\includegraphics[width=0.174\textwidth,height=0.115\textheight]{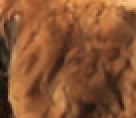} &
					\hspace{-3mm}
					\includegraphics[width=0.174\textwidth,height=0.115\textheight]{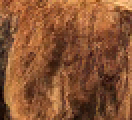} &
					\hspace{-3mm}
					\includegraphics[width=0.174\textwidth,height=0.115\textheight]{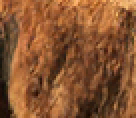} &
					\hspace{-3mm}
					\includegraphics[width=0.174\textwidth,height=0.115\textheight]{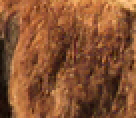} \\
					
					HR & \hspace{-3mm} ENet~\cite{EnhanceNet} & \hspace{-3mm} CX~\cite{CX} & \hspace{-3mm} EPSR3~\cite{EPSR} & 
					SuperSR~\cite{ESRGAN} & \hspace{-3mm} ESRGAN~\cite{ESRGAN} & \hspace{-3mm} PPON\_128 (Ours) & \hspace{-3mm} PPON (Ours) \\
					
					PSNR/LPIPS & \hspace{-3mm} 24.12/0.1795 & \hspace{-3mm} 25.03/0.1697 & \hspace{-3mm} 22.55/0.3107 &
					24.07/0.1632 & \hspace{-3mm} 22.73/0.2307 & \hspace{-3mm} 24.37/\textbf{0.1461} & \hspace{-3mm} 23.99/0.1547 \\
				\end{tabular}
			\end{adjustbox}
			\\
			\begin{adjustbox}{valign=t}
				\scriptsize
				\begin{tabular}{c}
					\includegraphics[width=0.2\textwidth, height=0.116\textheight]{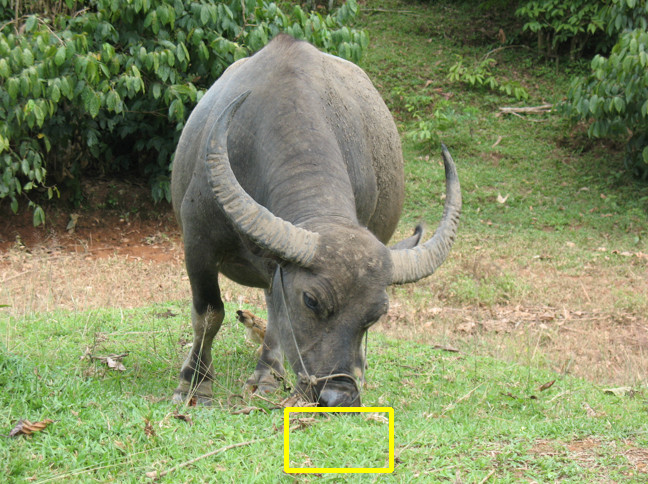} \\
					PIRM\_Test ($4 \times$): \\
					248 \\
				\end{tabular}
			\end{adjustbox}
			\hspace{-3mm}
			\begin{adjustbox}{valign=t}
				\scriptsize
				\begin{tabular}{cccccccc}
					\includegraphics[width=0.174\textwidth,height=0.115\textheight]{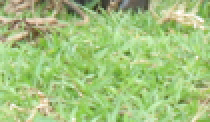} &
					\hspace{-3mm}
					\includegraphics[width=0.174\textwidth,height=0.115\textheight]{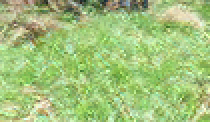} &
					\hspace{-3mm}
					\includegraphics[width=0.174\textwidth,height=0.115\textheight]{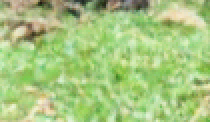} &
					\hspace{-3mm}
					\includegraphics[width=0.174\textwidth,height=0.115\textheight]{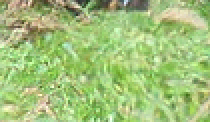} &
					\hspace{-3mm}					
					\includegraphics[width=0.174\textwidth,height=0.115\textheight]{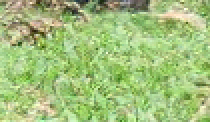} &
					\hspace{-3mm}
					\includegraphics[width=0.174\textwidth,height=0.115\textheight]{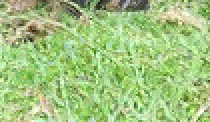} &
					\hspace{-3mm}
					\includegraphics[width=0.174\textwidth,height=0.115\textheight]{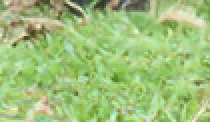} &
					\hspace{-3mm}
					\includegraphics[width=0.174\textwidth,height=0.115\textheight]{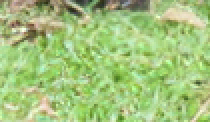} \\
					
					HR & \hspace{-3mm} ENet~\cite{EnhanceNet} & \hspace{-3mm} CX~\cite{CX} & \hspace{-3mm} EPSR3~\cite{EPSR} & 
					SuperSR~\cite{ESRGAN} & \hspace{-3mm} ESRGAN~\cite{ESRGAN} & \hspace{-3mm} PPON\_128 (Ours) & \hspace{-3mm} PPON (Ours) \\
					
					PSNR/LPIPS & \hspace{-3mm} 23.82/0.2265 & \hspace{-3mm} 24.83/0.1891 & \hspace{-3mm} 23.80/0.2250 &
					23.06/0.2323 & \hspace{-3mm} 22.55/0.2782 & \hspace{-3mm} 24.62/0.1593 & \hspace{-3mm} 24.26/\textbf{0.1544} \\
				\end{tabular}
			\end{adjustbox}
			\\
			\begin{adjustbox}{valign=t}
				\scriptsize
				\begin{tabular}{c}
					\includegraphics[width=0.2\textwidth, height=0.116\textheight]{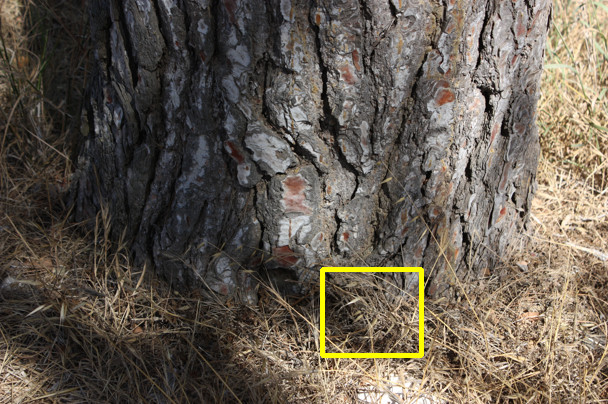} \\
					PIRM\_Test ($4 \times$): \\
					258 \\
				\end{tabular}
			\end{adjustbox}
			\hspace{-3mm}
			\begin{adjustbox}{valign=t}
				\scriptsize
				\begin{tabular}{cccccccc}
					\includegraphics[width=0.174\textwidth,height=0.115\textheight]{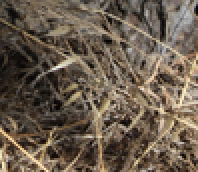} &
					\hspace{-3mm}
					\includegraphics[width=0.174\textwidth,height=0.115\textheight]{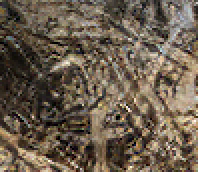} &
					\hspace{-3mm}
					\includegraphics[width=0.174\textwidth,height=0.115\textheight]{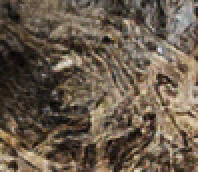} &
					\hspace{-3mm}
					\includegraphics[width=0.174\textwidth,height=0.115\textheight]{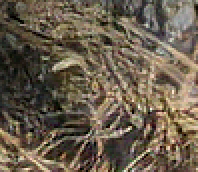} &
					\hspace{-3mm}					
					\includegraphics[width=0.174\textwidth,height=0.115\textheight]{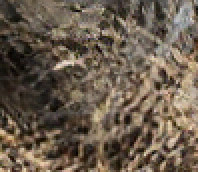} &
					\hspace{-3mm}
					\includegraphics[width=0.174\textwidth,height=0.115\textheight]{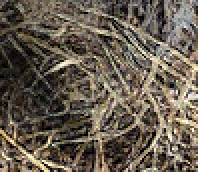} &
					\hspace{-3mm}
					\includegraphics[width=0.174\textwidth,height=0.115\textheight]{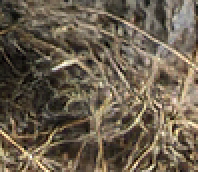} &
					\hspace{-3mm}
					\includegraphics[width=0.174\textwidth,height=0.115\textheight]{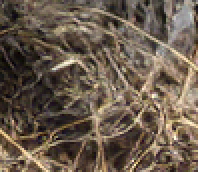} \\
					
					HR & \hspace{-3mm} ENet~\cite{EnhanceNet} & \hspace{-3mm} CX~\cite{CX} & \hspace{-3mm} EPSR3~\cite{EPSR} & 
					SuperSR~\cite{ESRGAN} & \hspace{-3mm} ESRGAN~\cite{ESRGAN} & \hspace{-3mm} PPON\_128 (Ours) & \hspace{-3mm} PPON (Ours) \\
					
					PSNR/LPIPS & \hspace{-3mm} 19.92/0.2776 & \hspace{-3mm} 20.41/0.2504 & \hspace{-3mm} 20.27/0.2938 &
					20.54/0.2616 & \hspace{-3mm} 18.67/0.2741 & \hspace{-3mm} 20.17/0.2202 & \hspace{-3mm} 20.43/\textbf{0.2068} \\
				\end{tabular}
			\end{adjustbox}
			\\
	\end{tabular} }
	\caption{Qualitative comparisons of perceptual-driven SR methods with our results at scaling factor of 4. Here, SuperSR is the variant of ESRGAN and it won the first place in the PIRM2018-SR Challenge.}
	\label{fig:compare-Pecetual}
\end{figure*}

Table~\ref{tab:pi} shows our quantitative evaluation results compared with $6$ perceptual-driven state-of-the-arts approaches: SRGAN~\cite{SRGAN}, ENet~\cite{EnhanceNet}, CX~\cite{CX}, EPSR~\cite{EPSR}, NatSR~\cite{NatSR}, and ESRGAN~\cite{ESRGAN}. The proposed PPON achieves the best in terms of LPIPS and keep the presentable PSNR values. For image ``86'' in Figures~\ref{fig:compare-Pecetual}, the result generated by S-RFN is blurred but has a elegant structure. Based on S-RFN, our PPON can synthesize realistic textures while retaining a delicate structure. It also validates the effectiveness of the proposed progressive architecture.

\begin{table}[htpb]
	\caption{Quantitative results about noise image super-resolution. RNAN\_DN is the RGB image denoising version of RNAN. Similarly, RNAN\_SR is the RGB image super-resolution version of RNAN. Noise level $\sigma  = 10$. The best results are \textbf{highlighted}.}
	\label{tab:noise10}
	\begin{center}
		\scalebox{0.92}{
		\begin{tabular}{|l|c|c|c|}
			\hline
			\multirow{2}{*}{Dataset} & RNAN\_DN + RNAN\_SR~\cite{RNAN} & RFN & S-RFN \\
			\cline{2-4}
			& PSNR / SSIM & PSNR / SSIM & PSNR / SSIM \\
			\hline
			\hline
			Set5~\cite{Set5} & 29.72 / 0.8693 & \textbf{30.17} / 0.8784 &  30.15 / \textbf{0.8790} \\
			
			Set14~\cite{Set14} & 27.30 / 0.7330 & \textbf{27.50} / 0.7395 & 27.48 / \textbf{0.7424} \\
			
			BSD100~\cite{BSD100} & 26.49 / 0.6827 &  \textbf{26.62} / 0.6877 & 26.60 / \textbf{0.6917} \\
			
			Urban100~\cite{Urban100} & 24.88 / 0.7354 & \textbf{25.47} / 0.7581 & 25.45 / \textbf{0.7600} \\
			
			Manga109~\cite{Manga109} & 28.41 / 0.8661 & \textbf{28.98} / 0.8802 & 28.96 / \textbf{0.8810} \\
			
			PIRM\_Val~\cite{PIRM-SR} & 27.07 / 0.7154 & \textbf{27.20} / 0.7217 & 27.17 / \textbf{0.7253} \\
			
			PIRM\_Test~\cite{PIRM-SR} & 27.04 / 0.7048 & \textbf{27.15} / 0.7103 & 27.13 / \textbf{0.7141} \\
			
			\hline
		\end{tabular} }
	\end{center}
\end{table}

\begin{table}[htpb]
	\caption{Average resolution/time evaluated on seven datasets (JPEG LR $\times 4$ SR).}
	\label{tab:time}
	\begin{center}
		\scalebox{0.92}{
			\begin{tabular}{|l|c|c|c|}
				\hline
				Dataset & Input resolution (px, $H \times W$) & Memory (MB) & Time (ms) \\
				\hline
				\multirow{2}{*}{PIRM\_Test} & $121 \times 152$ & 1,171 & 206 \\
				\cline{2-4}
				& $242 \times 305$ & 4,087 & 745 \\
				\hline
				
				\multirow{2}{*}{PIRM\_Val} & $119 \times 155$ & 1,267 & 213 \\
				\cline{2-4}
				& $239 \times 311$ & 2,495 & 759 \\
				\hline
				
				\multirow{2}{*}{Set5} & $84 \times 72$ & 899 & 107 \\
				\cline{2-4}
				& $168 \times 156$ & 1,607 & 305 \\
				\hline
				
				\multirow{2}{*}{Set14} & $111 \times 122$ & 1399 & 163 \\
				\cline{2-4}
				& $222 \times 245$ & 2,089 & 577 \\
				\hline
				
				\multirow{2}{*}{B100} & $89 \times 111$ & 809 & 111 \\
				\cline{2-4}
				& $178 \times 221$ & 1,211 & 401 \\
				\hline
				
				\multirow{2}{*}{Urban100} & $199 \times 246$ & 2,047 & 501 \\
				\cline{2-4}
				& $398 \times 492$ & 6,583 & 2,028 \\
				\hline
				
				\multirow{2}{*}{Manga109} & $291 \times 205$ & 1,539 & 628 \\
				\cline{2-4}
				& $584 \times 412$ & 3,923 & 2,580 \\
				\hline
				
				\hline
		\end{tabular} }
	\end{center}
\end{table}

\begin{figure*}[htpb]
	\centering
	\scalebox{0.7}{
		\begin{tabular}{lc}
			% 42049 BSD100
			\begin{adjustbox}{valign=t}
				\scriptsize
				\begin{tabular}{c}
					\includegraphics[width=0.2\textwidth, height=0.127\textheight]{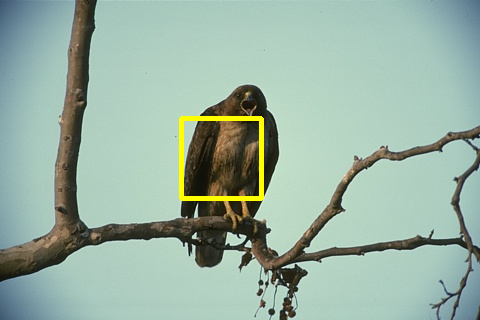} \\
					BSD100 ($4 \times$): \\
					42049 \\
				\end{tabular}
			\end{adjustbox}
			\hspace{-3mm}
			\begin{adjustbox}{valign=t}
				\begin{tabular}{cccccc}
					\includegraphics[width=0.174\textwidth, height=0.126\textheight]{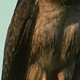} &
					\hspace{-3mm}
					\includegraphics[width=0.174\textwidth, height=0.126\textheight]{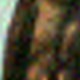} &
					\hspace{-3mm}
					\includegraphics[width=0.174\textwidth, height=0.126\textheight]{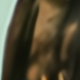} &
					\hspace{-3mm}
					\includegraphics[width=0.174\textwidth, height=0.126\textheight]{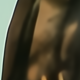} &
					\hspace{-3mm}
					\includegraphics[width=0.174\textwidth, height=0.126\textheight]{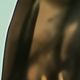} &
					\hspace{-3mm}
					\includegraphics[width=0.174\textwidth, height=0.126\textheight]{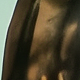} \\
					
					HR & \hspace{-3mm} Noisy ($\sigma  = 10$) & \hspace{-3mm} RNAN~\cite{RNAN} & \hspace{-3mm} RFN & \hspace{-3mm} S-RFN & \hspace{-3mm} PPON \\
				\end{tabular}
			\end{adjustbox}
			\\
			
			\begin{adjustbox}{valign=t}
				\scriptsize
				\begin{tabular}{c}
					\includegraphics[width=0.2\textwidth, height=0.127\textheight]{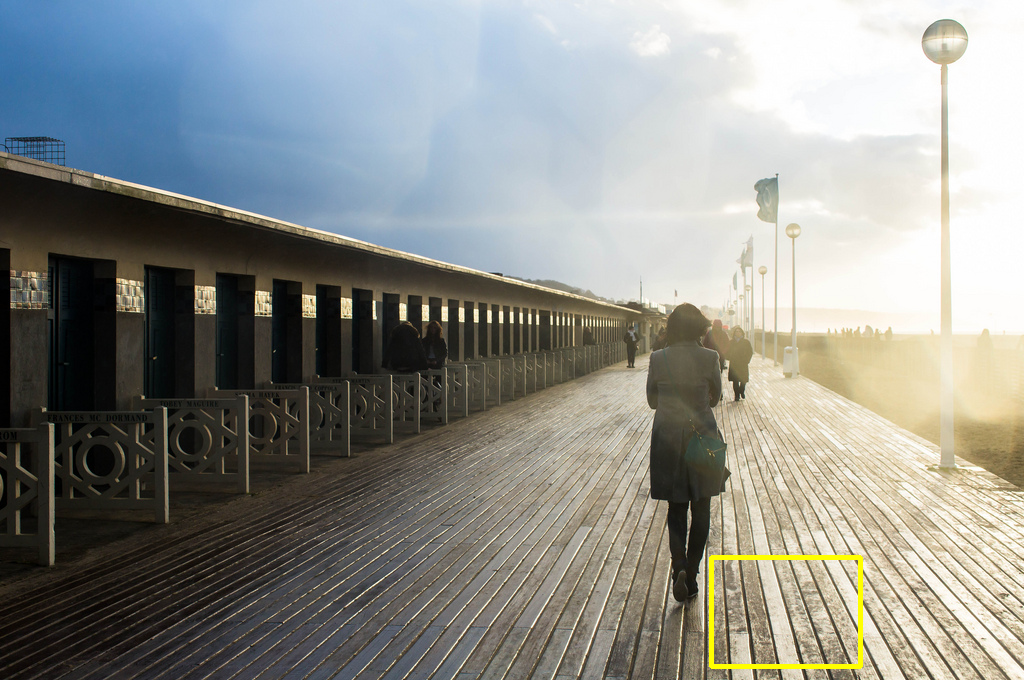} \\
					Urban100 ($4 \times$): \\
					img\_032 \\
				\end{tabular}
			\end{adjustbox}
			\hspace{-3mm}
			\begin{adjustbox}{valign=t}
				\begin{tabular}{cccccc}
					\includegraphics[width=0.174\textwidth,height=0.126\textheight]{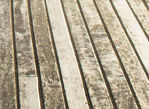} &
					\hspace{-3mm}
					\includegraphics[width=0.174\textwidth,height=0.126\textheight]{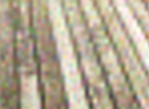} &
					\hspace{-3mm}
					\includegraphics[width=0.174\textwidth,height=0.126\textheight]{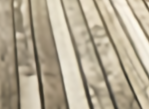} &
					\hspace{-3mm}
					\includegraphics[width=0.174\textwidth,height=0.126\textheight]{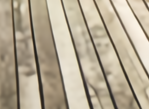} &
					\hspace{-3mm}
					\includegraphics[width=0.174\textwidth,height=0.126\textheight]{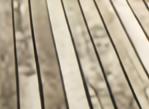} &
					\hspace{-3mm}
					\includegraphics[width=0.174\textwidth,height=0.126\textheight]{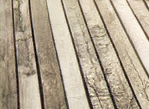} \\
					
					HR & \hspace{-3mm} Noisy ($\sigma  = 10$) & \hspace{-3mm} RNAN~\cite{RNAN} & \hspace{-3mm} RFN & \hspace{-3mm} S-RFN & \hspace{-3mm} PPON \\
				\end{tabular}
			\end{adjustbox}
			\\
	\end{tabular} }
	\caption{Noise image super-resolution results with noise level $\sigma  = 10$.}
	\label{fig:noise10}
\end{figure*}
We further apply our PPON to solve the noise image super-resolution. AWGN noises (noise level is set to 10) are added to clean low-resolution images. Quantitative results are shown in Table~\ref{tab:noise10}. It is noted that we only fine-tune the COBranch by noise training images and maintain the SOBranch and POBranch. In this way, the produced structure-aware and perceptual-aware results are still steady as we can see that our RFN achieves the best PSNR performance, and S-RFN achieves the best SSIM performance, which is consistent with the results in Table~\ref{tab:psnr-ssim-values}. Even if SOBranch does not retrain by noise-clean images pairs, S-RFN still obtains higher SSIM scores than RFN. It also suggests that the separability of PPON. We also show visual results in Figure~\ref{fig:noise10}. Obviously, RFN and S-RFN can generate sharper edges (``42049'' from BSD100 and ``img\_032'' from Urban100), and PPON can hallucinate some plausible details.

\begin{figure*}[htpb]
	\centering
	\scalebox{0.82}{
		\begin{tabular}{lc}
			% HealingPlanet Manga109
			\begin{adjustbox}{valign=t}
				\scriptsize
				\begin{tabular}{c}
					\includegraphics[width=0.2\textwidth, height=0.127\textheight]{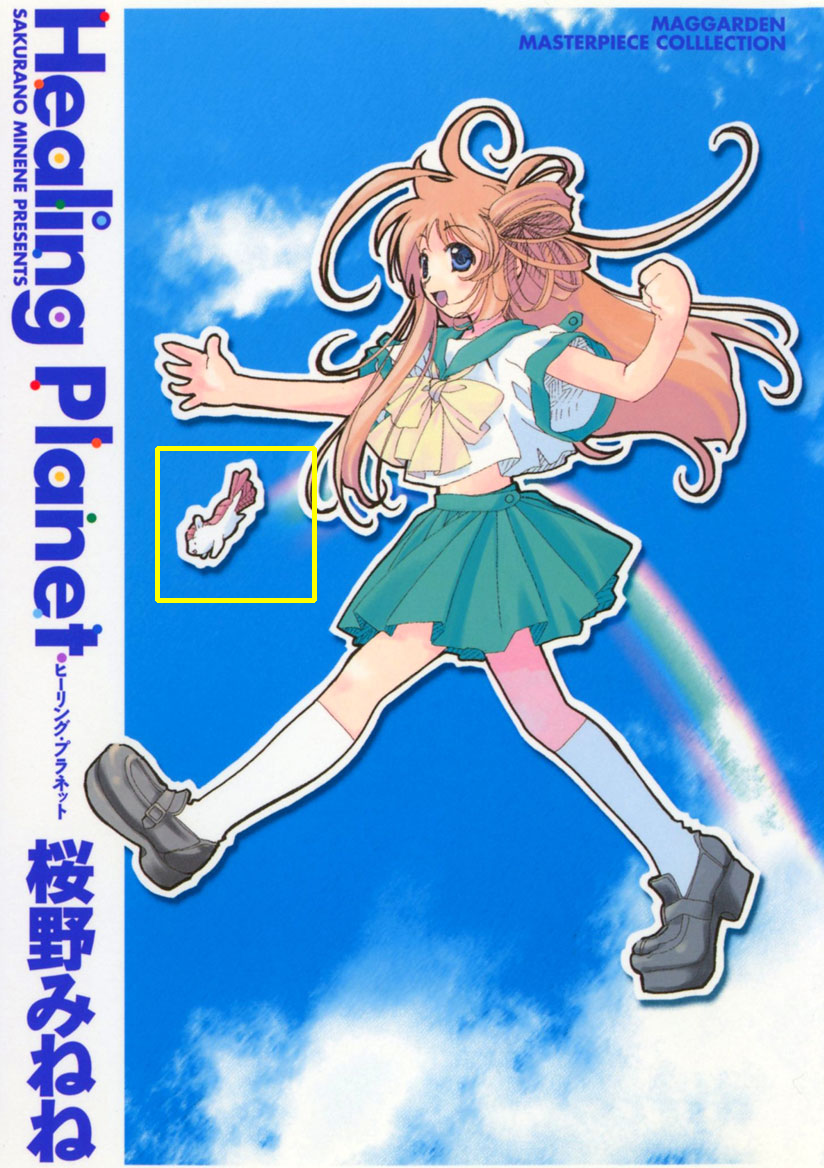} \\
					Manga109 ($4 \times$): \\
					HealingPlanet \\
				\end{tabular}
			\end{adjustbox}
			\hspace{-3mm}
			\begin{adjustbox}{valign=t}
				\begin{tabular}{ccccc}
					\includegraphics[width=0.174\textwidth, height=0.126\textheight]{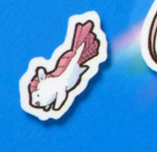} &
					\hspace{-3mm}
					\includegraphics[width=0.174\textwidth, height=0.126\textheight]{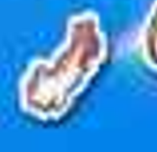} &
					\hspace{-3mm}
					\includegraphics[width=0.174\textwidth, height=0.126\textheight]{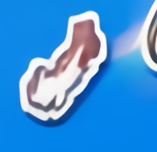} &
					\hspace{-3mm}
					\includegraphics[width=0.174\textwidth, height=0.126\textheight]{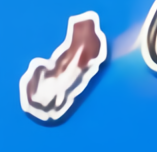} &
					\hspace{-3mm}
					\includegraphics[width=0.174\textwidth, height=0.126\textheight]{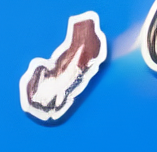} \\
					
					HR & \hspace{-3mm} JPEG ($q  = 40$) & \hspace{-3mm} RFN & \hspace{-3mm} S-RFN & \hspace{-3mm} PPON \\
				\end{tabular}
			\end{adjustbox}
			\\
			
			\begin{adjustbox}{valign=t}
				\scriptsize
				\begin{tabular}{c}
					\includegraphics[width=0.2\textwidth, height=0.127\textheight]{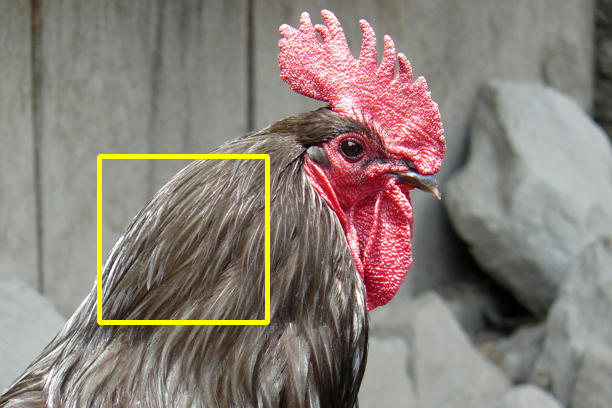} \\
					PIRM\_Test ($4 \times$): \\
					242 \\
				\end{tabular}
			\end{adjustbox}
			\hspace{-3mm}
			\begin{adjustbox}{valign=t}
				\begin{tabular}{ccccc}
					\includegraphics[width=0.174\textwidth,height=0.126\textheight]{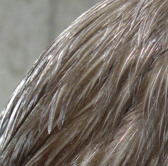} &
					\hspace{-3mm}
					\includegraphics[width=0.174\textwidth,height=0.126\textheight]{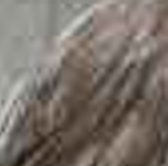} &
					\hspace{-3mm}
					\includegraphics[width=0.174\textwidth,height=0.126\textheight]{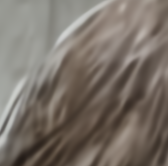} &
					\hspace{-3mm}
					\includegraphics[width=0.174\textwidth,height=0.126\textheight]{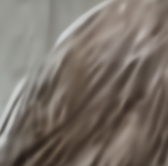} &
					\hspace{-3mm}
					\includegraphics[width=0.174\textwidth,height=0.126\textheight]{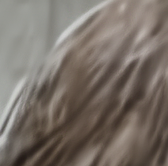} \\
					
					HR & \hspace{-3mm} JPEG ($q  = 40$) & \hspace{-3mm} RFN & \hspace{-3mm} S-RFN & \hspace{-3mm} PPON \\
				\end{tabular}
			\end{adjustbox}
			\\
	\end{tabular} }
	\caption{JPEG compressed image super-resolution results with JPEG quality $q  = 40$.}
	\label{fig:jpeg40}
\end{figure*}

We further apply our PPON to upscale LR images with compression artifacts. Due to the previous image compression artifacts methods focusing on the restoration of the Y channel (in YCbCr space), we only show our visual results in Figure~\ref{fig:jpeg40} (RGB JPEG compression artifacts reduction and super-resolution). From Figure~\ref{fig:jpeg40}, we can observe that our method can process the low-quality input well (clean edge, clean background).
To probe into the influence of resolution of the input images with JPEG compression, we feed JPEG compressed LR images with different spatial resolutions into our PPON, and then we explore memory 
occupation and inference on seven datasets (see in Table~\ref{tab:time}). If the input resolution increased to twice, the memory and time consumption increased to less than 4 times. It suggests our model can run on large resolution image well, considering memory and speed.

\begin{figure*}[htpb]
	\centering
		\begin{tabular}{lc}
			% img_091 Urban100
			\begin{adjustbox}{valign=t}
				\scriptsize
				\begin{tabular}{c}
					\includegraphics[width=0.2\textwidth, height=0.127\textheight]{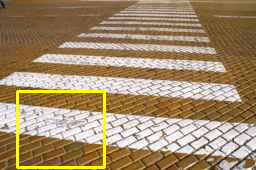} \\
					img\_091: \\
					($170 \times 256$) \\
				\end{tabular}
			\end{adjustbox}
			\hspace{-3mm}
			\begin{adjustbox}{valign=t}
				\begin{tabular}{cccc}
					\includegraphics[width=0.174\textwidth, height=0.126\textheight]{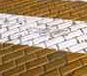} &
					\hspace{-3mm}
					\includegraphics[width=0.174\textwidth, height=0.126\textheight]{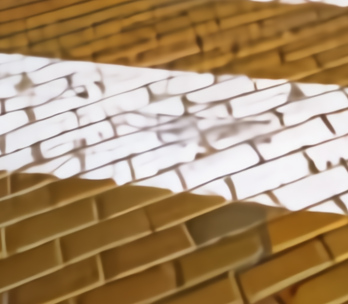} &
					\hspace{-3mm}
					\includegraphics[width=0.174\textwidth, height=0.126\textheight]{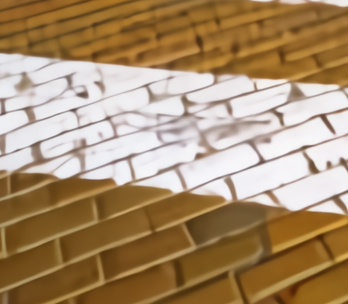} &
					\hspace{-3mm}
					\includegraphics[width=0.174\textwidth, height=0.126\textheight]{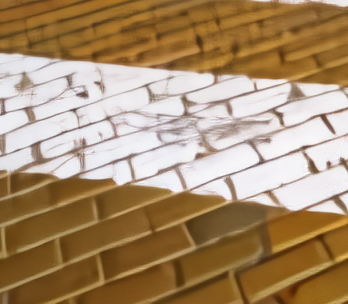} \\

					LR ($q  = 40$) & \hspace{-3mm} RFN  & \hspace{-3mm} S-RFN & \hspace{-3mm} PPON \\
				\end{tabular}
			\end{adjustbox}
			\\
			
			\begin{adjustbox}{valign=t}
				\scriptsize
				\begin{tabular}{c}
					\includegraphics[width=0.2\textwidth, height=0.127\textheight]{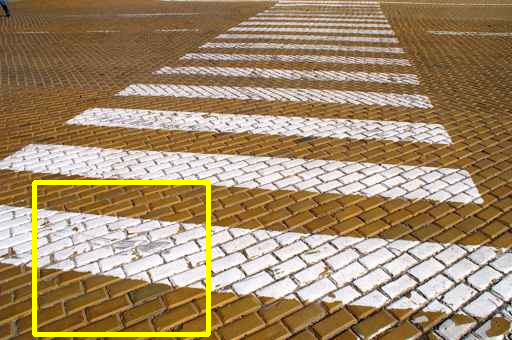} \\
					img\_091: \\
					($340 \times 512$) \\
				\end{tabular}
			\end{adjustbox}
			\hspace{-3mm}
			\begin{adjustbox}{valign=t}
				\begin{tabular}{cccc}
					\includegraphics[width=0.174\textwidth,height=0.126\textheight]{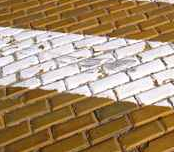} &
					\hspace{-3mm}
					\includegraphics[width=0.174\textwidth,height=0.126\textheight]{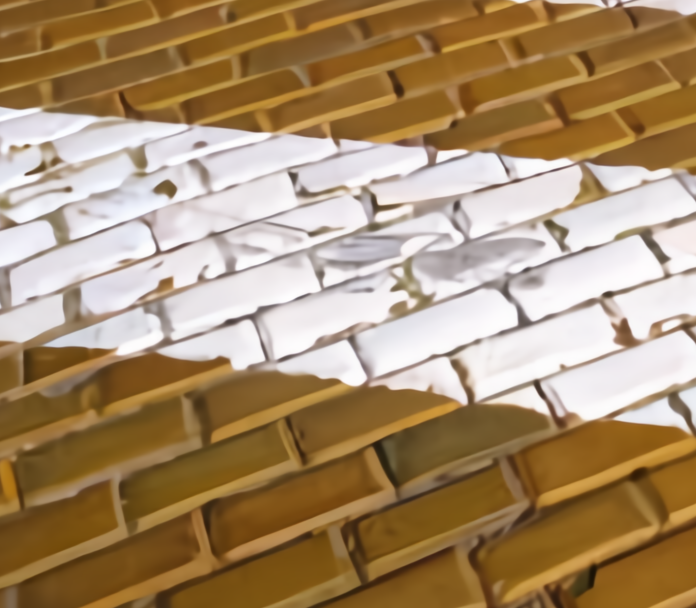} &
					\hspace{-3mm}
					\includegraphics[width=0.174\textwidth,height=0.126\textheight]{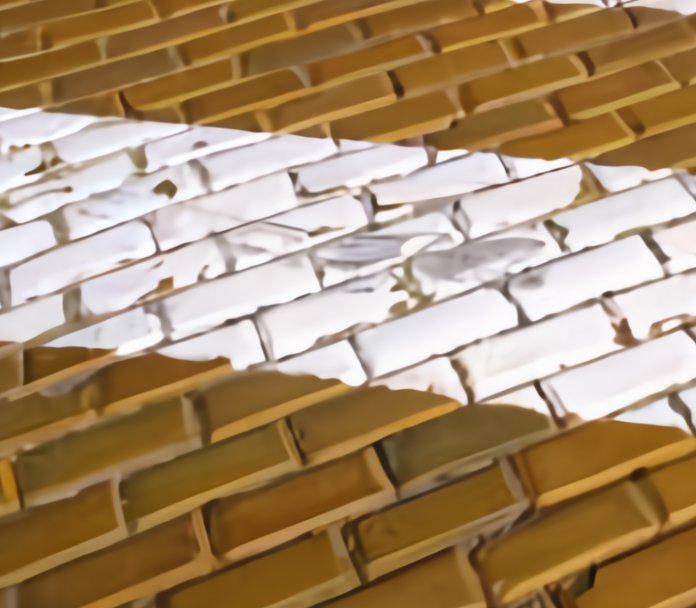} &
					\hspace{-3mm}
					\includegraphics[width=0.174\textwidth,height=0.126\textheight]{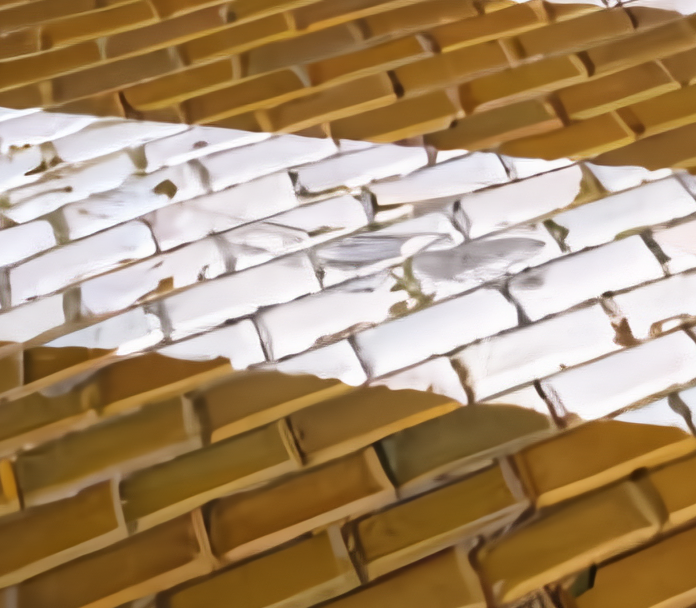} \\
					
					LR ($q  = 40$) & \hspace{-3mm} RFN & \hspace{-3mm} S-RFN & \hspace{-3mm} PPON \\
				\end{tabular}
			\end{adjustbox}
			\\
			% TennenSenshiG Manga109
			\begin{adjustbox}{valign=t}
				\scriptsize
				\begin{tabular}{c}
					\includegraphics[width=0.2\textwidth, height=0.127\textheight]{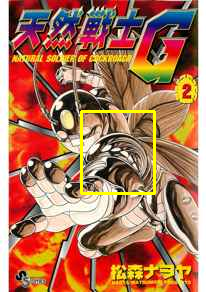} \\
					TennenSenshiG: \\
					($292 \times 206$) \\
				\end{tabular}
			\end{adjustbox}
			\hspace{-3mm}
			\begin{adjustbox}{valign=t}
				\begin{tabular}{cccc}
					\includegraphics[width=0.174\textwidth, height=0.126\textheight]{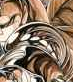} &
					\hspace{-3mm}
					\includegraphics[width=0.174\textwidth, height=0.126\textheight]{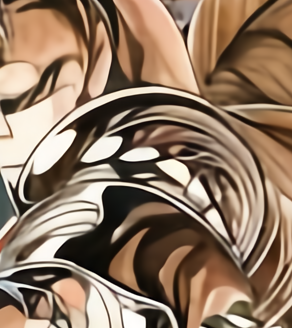} &
					\hspace{-3mm}
					\includegraphics[width=0.174\textwidth, height=0.126\textheight]{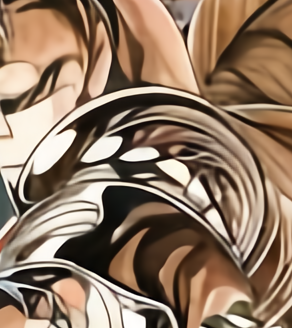} &
					\hspace{-3mm}
					\includegraphics[width=0.174\textwidth, height=0.126\textheight]{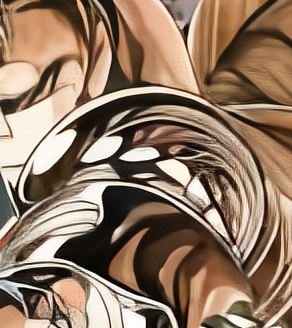} \\

					LR ($q  = 40$) & \hspace{-3mm} RFN  & \hspace{-3mm} S-RFN & \hspace{-3mm} PPON \\
				\end{tabular}
			\end{adjustbox}
			\\
			\begin{adjustbox}{valign=t}
				\scriptsize
				\begin{tabular}{c}
					\includegraphics[width=0.2\textwidth, height=0.127\textheight]{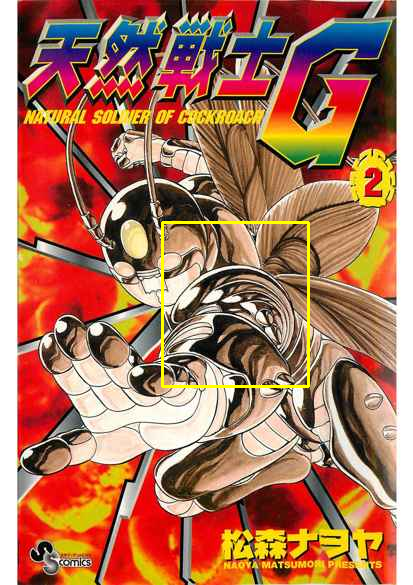} \\
					TennenSenshiG: \\
					($585 \times 413$) \\
				\end{tabular}
			\end{adjustbox}
			\hspace{-3mm}
			\begin{adjustbox}{valign=t}
				\begin{tabular}{cccc}
					\includegraphics[width=0.174\textwidth,height=0.126\textheight]{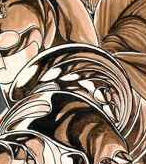} &
					\hspace{-3mm}
					\includegraphics[width=0.174\textwidth,height=0.126\textheight]{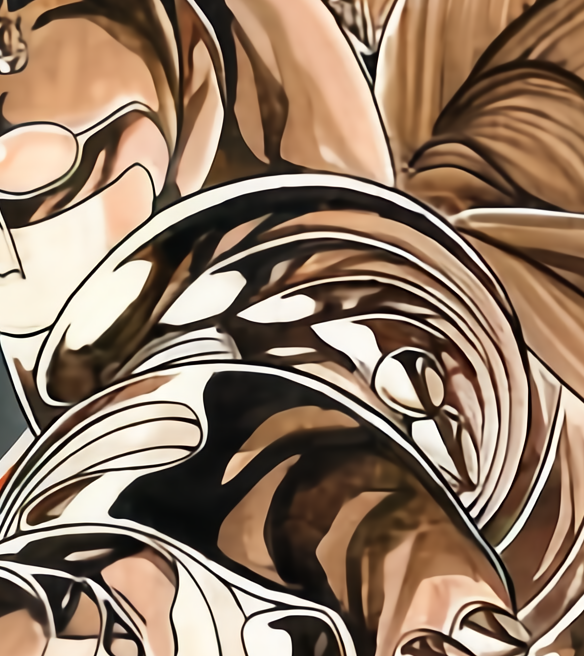} &
					\hspace{-3mm}
					\includegraphics[width=0.174\textwidth,height=0.126\textheight]{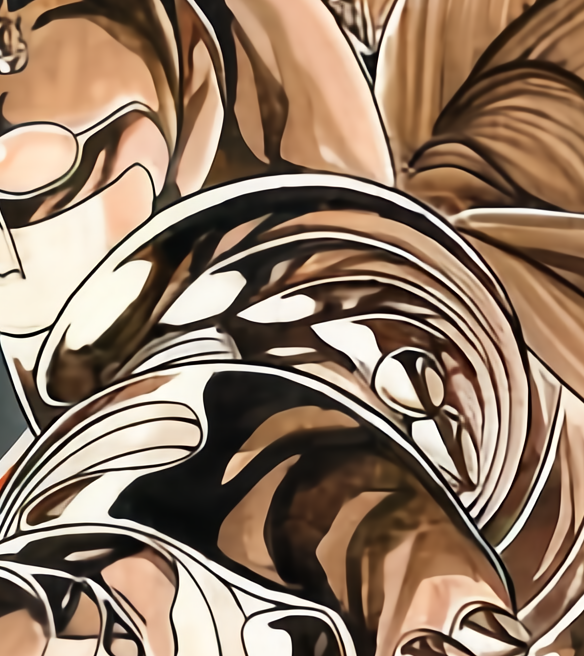} &
					\hspace{-3mm}
					\includegraphics[width=0.174\textwidth,height=0.126\textheight]{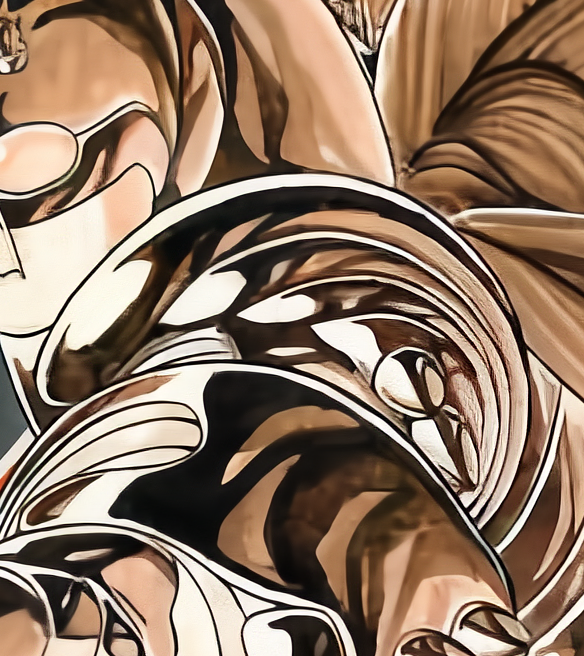} \\
					
					LR ($q  = 40$) & \hspace{-3mm} RFN & \hspace{-3mm} S-RFN & \hspace{-3mm} PPON \\
				\end{tabular}
			\end{adjustbox}
			\\
	\end{tabular}
	\caption{JPEG compressed image super-resolution results with JPEG quality $q  = 40$ and different input resolutions. Here, two qualitative results from Urban100 and Manga109, respectively.}
	\label{fig:jpeg40-resolution}
\end{figure*}

In Figure~\ref{fig:jpeg40-resolution}, two qualitative results are showed to verify that the high-resolution input image does gain better super-resolved images. For example, ``img\_091'' with the spatial resolution $170 \times 256$ is low quality, the generated images from RFN and S-RFN are similar, and PPON produces an image that is slightly better effect. When the input resolution is increasing to $340 \times 512$, three results (RFN, S-RFN, and PPON) are of high quality. It demonstrates that our model can handle low-resolution images and high-resolution images: better quality input and better quality output.

\subsection{The choice of main evaluation metric}\label{sec:Metric}
\begin{figure*}[htpb]
	\scriptsize
	\centering
	\scalebox{0.58}{
	\begin{tabular}{lc}
		\begin{adjustbox}{valign=t}
			\begin{tabular}{c}
				\includegraphics[width=0.35\textwidth,height=0.3\textheight]{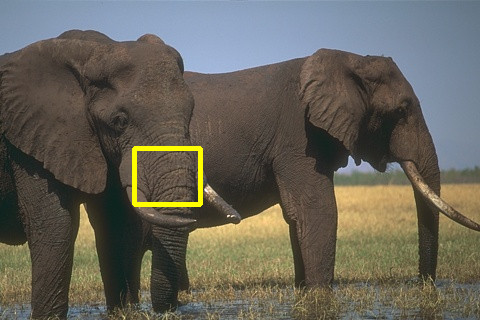} \\
				296059 from BSD100 \\
				(PSNR / LPIPS / PI) \\
			\end{tabular}
		\end{adjustbox}
		\hspace{-3mm}
		\begin{adjustbox}{valign=t}
			\scriptsize
			\begin{tabular}{ccccc}
				\includegraphics[width=0.25\textwidth, height=0.125\textheight]{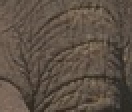} &
				\hspace{-3mm}
				\includegraphics[width=0.25\textwidth, height=0.125\textheight]{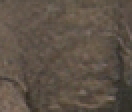} &
				\hspace{-3mm}
				\includegraphics[width=0.25\textwidth, height=0.125\textheight]{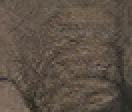} &
				\hspace{-3mm}
				\includegraphics[width=0.25\textwidth, height=0.125\textheight]{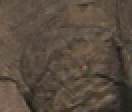} &
				\hspace{-3mm}
				\includegraphics[width=0.25\textwidth, height=0.125\textheight]{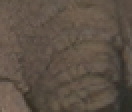}  \\
				HR & SRGAN~\cite{SRGAN} & \hspace{-3mm} ENet~\cite{EnhanceNet} & \hspace{-3mm} CX~\cite{CX} & \hspace{-3mm} EPSR2~\cite{EPSR} \\
				($\infty $ / 0 / 2.3885) & \hspace{-3mm} (28.96 / 0.1564 / 2.6015) & \hspace{-3mm} (29.18 / 0.1432 / 2.8138) & \hspace{-3mm} (28.57 / 0.1563 / 2.3492) & \hspace{-3mm} (30.47 / 0.2046 /3.2575) \\
				\includegraphics[width=0.25\textwidth, height=0.125\textheight]{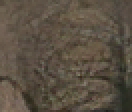} &
				\hspace{-3mm}
				\includegraphics[width=0.25\textwidth, height=0.125\textheight]{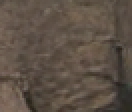} &
				\hspace{-3mm}
				\includegraphics[width=0.25\textwidth, height=0.125\textheight]{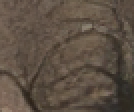} &
				\hspace{-3mm}
				\includegraphics[width=0.25\textwidth, height=0.125\textheight]{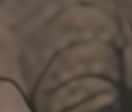} &
				\hspace{-3mm}
				\includegraphics[width=0.25\textwidth, height=0.125\textheight]{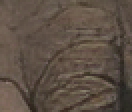}  \\
				EPSR3~\cite{EPSR} & SuperSR~\cite{ESRGAN} & ESRGAN~\cite{ESRGAN} & S-RFN (Ours) & PPON (Ours) \\
				(29.02 / 0.1911 /2.2666) & \hspace{-3mm} (29.80 / 0.1703 /2.2913) & \hspace{-3mm} (29.38 / 0.1333 / 2.3481) & \hspace{-3mm} (31.40 / 0.3314 / 4.7222) & \hspace{-3mm} (29.26 / \textbf{0.1305} / 2.5130)
			\end{tabular}
		\end{adjustbox}
	\end{tabular} }
	\caption{A visual comparison with the state-of-the-art perceptual image SR algorithms.}
	\label{fig:verify-metric}
\end{figure*}

We consider LPIPS\footnote{\url{https://github.com/richzhang/PerceptualSimilarity}}~\cite{LPIPS} and PI\footnote{\url{https://github.com/roimehrez/PIRM2018}}~\cite{PIRM-SR} as our evaluation indices of perceptual image SR. As illustrated in Figure~\ref{fig:verify-metric}, we can see that the PI score of EPSR3 (\textbf{2.2666}) is even better than HR (\textbf{2.3885}), but EPSR3 shows unnatural and lacks proper texture and structure. When observing the results of ESRGAN and our PPON, their perception effect is superior to that of EPSR3, precisely in accordance with corresponding LPIPS values. From the results of S-RFN and PPON, it can be demonstrated that both PI and LPIPS have the ability to distinguish a blurring image. From the images of EPSR3, SuperSR, and ground-truth (HR), we can distinctly know that the lower PI value does not mean better image quality. Compared with the image generated by ESRGAN~\cite{ESRGAN}, it is evident that the proposed PPON gets a better visual effect with more structure information, corresponding to the lower LPIPS value. Because the PI (non-reference measure) is not sensitive to deformation through the experiment and cannot reflect the similarity with ground-truth, we take LPIPS as our primary perceptual measure and PI as a secondary metric.

Besides, we performed a MOS (mean opinion score) test to validate the effectiveness of our PPON further. Specifically, we collect $16$ raters to assign an integral score from $1$ (bad quality) to $5$ (excellent quality). To ensure the reliability of the results, we provide the raters with tests and original HR images simultaneously. The ground-truth images are set to $5$, and the raters then score the test images based on it. The average MOS results are shown in Table~\ref{tab:MOS}. 
%---------------------------
% MOS
%---------------------------
\begin{table}[htpb]
	\centering
	\caption{Comparison of CX, ESRGAN, S-RFN, and PPON.}
	\label{tab:MOS}
	\begin{tabular}{|c|c|c|c|c|}
		\hline
		PIRM\_Val & CX & ESRGAN & S-RFN(Ours) & PPON(Ours) \\
		\hline
		MOS & 2.42 & \underline{3.23} & 1.82 & \textbf{3.58} \\
		PSNR & 25.41 & 25.18 & \textbf{28.63} & \underline{26.20} \\
		SSIM & 0.6747 & 0.6596 & \textbf{0.7913} & \underline{0.6995} \\
		\hline
	\end{tabular}
\end{table}

\subsection{The influence of training patch size}
\begin{table}[htpb]
	\caption{Quantitative evaluation of different perceptual-driven SR methods in LPIPS and PI. PPON\_128 indicates the POBranch trained with $128 \times 128$ image patches. The best and second best results are \textbf{highlighted} and \underline{underlined}, respectively.}
	\label{tab:ppon-128-192}
	\begin{center}
		\begin{tabular}{|c|c|c|}
			\hline
			\multirow{2}{*}{Method} & PIRM\_Val & PIRM\_Test \\
			\cline{2-3}
			& LPIPS / PI & LPIPS/ PI \\
			\hline
			\hline
			ESRGAN~\cite{ESRGAN} & 0.1443 / 2.5550 & 0.1523 / 2.4356 \\
			PPON\_128 (Ours) & \underline{ 0.1241} / \underline{2.3026} & \underline{0.1321} / \underline{2.2080} \\
			PPON (Ours) & \textbf{0.1194} / \textbf{2.2736} & \textbf{0.1273} / \textbf{2.1770} \\
			\hline
		\end{tabular}
	\end{center}
\end{table}
In ESRGAN~\cite{ESRGAN}, the authors mentioned that larger training patch sizes cost more training time and consume more computing resources. Thus, they used $192 \times 192$ for PSNR-oriented methods and $128 \times 128$ for perceptual-driven methods. In our main manuscript, we train the COBranch, SOBranch, and POBranch with $192 \times 192$ image patches. Here, we further explore the influence of larger patches in the perceptual image generation stage.

\begin{figure*}[htpb]
	\centering
	\subfigure[Discriminator for $128 \times 128$ training patches in PPON\_128.]{
		\includegraphics[width=0.98\textwidth]{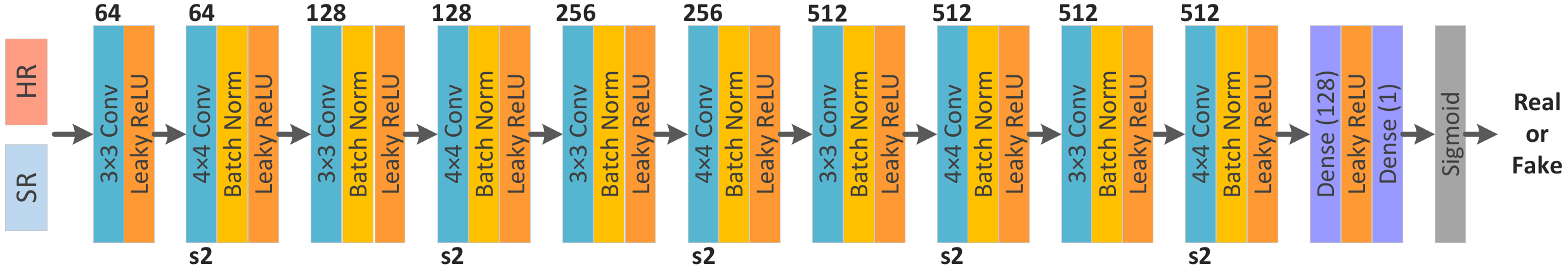}}
	\hfil
	\subfigure[Discriminator for $192 \times 192$ training patches in PPON.]{\includegraphics[width=0.98\textwidth]{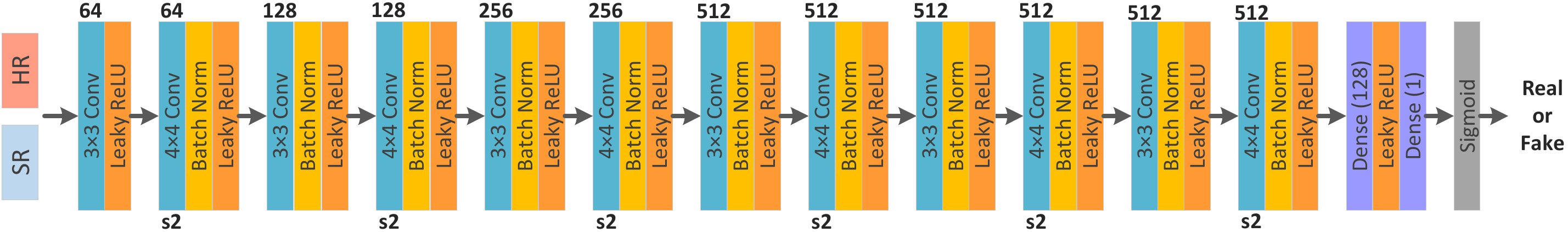}}
	\caption{The network structure of the discriminators. The output size is scaled down by stride 2, and the parameter of LReLU is $0.2$.}
	\label{fig:discriminator}
\end{figure*}

It is important to note that the training perceptual-driven model requires more GPU memory and ore considerable computing resources than the PSNR-oriented model since the VGG model and discriminator need to be loaded during the training of the former. Therefore, larger patches ($192 \times 192$) are hard to be used in optimizing ESRGAN~\cite{ESRGAN} due to their large generator and discriminator to be updated. Thanks to our POBranch containing very few parameters, we employ $192 \times 192$ training patches and achieve better results, as shown in Table~\ref{tab:ppon-128-192}. Concerning the discriminators, we illustrate them in Figure~\ref{fig:discriminator}. For a fair comparison with the ESRGAN~\cite{ESRGAN}, we retrain our POBranch with $128 \times 128$ patches and provide the results in Table~\ref{tab:ppon-128-192}. 

\section{Conclusion}\label{sec:conclusion}

In this paper, we propose a progressive perception-oriented network (PPON) for better perceptual image SR. Concretely, three branches are developed to learn the content, structure, and perceptual details, respectively. By exerting a stage-by-stage training scheme, we can steadily get promising results.  It is worth mentioning that these three branches are not independent. A structure-oriented branch can exploit the extracted features and output images of the content-oriented branch. Extensive experiments on both traditional SR and perceptual SR demonstrate the effectiveness of our proposed PPON. 

\section*{Acknowledgments}
This work was supported in part by the National Key Research and Development Program of China under Grants 2018AAA0102702, 2018AAA0103202, in part by the National Natural Science Foundation of China under Grant 61772402, 61671339, 61871308, and 61972305.

\section*{References}

\bibliography{mybibfile}

\end{document}